\documentclass[10pt,journal]{IEEEtran}
\usepackage{amsmath,amssymb,amsfonts}
\usepackage{amsthm}
\usepackage{bm}
\usepackage{graphicx}
\usepackage[caption=false,font=footnotesize]{subfig}
\usepackage{tikz}
\usetikzlibrary{arrows.meta, positioning, fit, backgrounds,
                shapes.geometric, decorations.pathreplacing, calc}
\definecolor{covblue}{RGB}{52,120,190}
\definecolor{absorborange}{RGB}{220,120,40}
\definecolor{infogreen}{RGB}{50,160,80}
\definecolor{srcred}{RGB}{210,60,50}
\usepackage{booktabs}
\usepackage{multirow}
\usepackage{algorithm}
\usepackage{algorithmic}
\usepackage{xcolor}
\newif\ifreviewmarkup \reviewmarkupfalse
\definecolor{revred}{RGB}{200,30,30}
\definecolor{revgreen}{RGB}{0,130,60}
\makeatletter
\DeclareRobustCommand{\was}{\dc@was}
\long\def\dc@was#1{\ifreviewmarkup{\color{revred}#1}\fi}
\makeatother
\makeatletter
\DeclareRobustCommand{\new}{\dc@new}
\long\def\dc@new#1{\ifreviewmarkup{\color{revgreen}#1}\else#1\fi}
\makeatother
\definecolor{revpend}{RGB}{200,120,0}
\makeatletter
\DeclareRobustCommand{\pending}{\dc@pend}
\long\def\dc@pend#1{{\color{revpend}[#1]}}
\makeatother
\definecolor{revexp}{RGB}{20,80,190}
\makeatletter
\DeclareRobustCommand{\expect}{\dc@exp}
\long\def\dc@exp#1{{\color{revexp}#1}}
\makeatother
\definecolor{revrun}{RGB}{20,80,190}

\usepackage{url}
\usepackage{cite}
\usepackage{placeins}
\usepackage{listings} 

\newcommand{\R}{\mathbb{R}}

\newcommand{\norm}[1]{\left\lVert#1\right\rVert}
\newtheorem{proposition}{Proposition}

\usepackage{xr-hyper}
\usepackage[colorlinks=true,urlcolor=blue,linkcolor=black,citecolor=black]{hyperref}
\usepackage{cleveref}
\crefname{equation}{Eq.}{Eqs.}
\crefname{figure}{Fig.}{Figs.}
\crefname{table}{Table}{Tables}
\crefname{section}{Sec.}{Secs.}
\crefname{algorithm}{Algorithm}{Algorithms}
\crefname{proposition}{Proposition}{Propositions}
\title{Gradient-Based Trajectory Optimisation over Continuous Poses for Sparse-View Cone-Beam CT}
\author{Linda-Sophie Schneider, Simon Wittl, Gabriel Herl and Andreas Maier%
\thanks{L.-S. Schneider and A. Maier are with the Pattern Recognition Lab, Friedrich-Alexander-Universit{\"a}t Erlangen-N{\"u}rnberg, Germany. Correspondence to linda-sophie.schneider@fau.de.}
\thanks{S. Wittl and G. Herl are with the Deggendorf Institute of Technology, Germany.}}
\begin{document}
\maketitle
\begin{abstract}
Trajectory optimisation for cone-beam computed tomography (CT) determines which information sparse-view scans acquire. Fixed candidate pools prevent off-grid refinement and require new object-specific precomputation for each acquisition manifold. We make every source pose an individual continuous variable and move all poses jointly by gradient ascent on the scanner's kinematic manifold. The objective combines soft-Tuy plane coverage, continuous View Covariance Loss, and an analytic attenuation-aware ray-bundle penalty. The same optimiser handles circular, limited C-arm, two-axis, and free-sphere parametrisations. On a Defrise flange, continuous selection recovers laminar defects invisible to a circular orbit, matches discrete swap search on the free sphere at the sparser budget, and leads at the denser one, with the same objective evaluated in every arm. A moderate elevation band already recovers most of the free-sphere gain at the defects, so the same optimiser transfers to bounded scanner envelopes. Photon noise preserves the ordering on the flange and compresses it on a dense fuel nozzle. Sparse-prescan planning benefits from matching prescan and planned acquisition manifolds. Selection takes seconds rather than minutes without an object-specific reconstruction basis. Prescan-planned poses were executed on a robot CT bench and reconstructed in a common frame, demonstrating feasibility but no consistent metric gain over uniform band sampling. Continuous pose optimisation incorporates attenuation and scanner constraints directly into sparse-view acquisition design.
\end{abstract}
\section{Introduction}
\label{sec:intro}
Trajectory optimisation for computed tomography (CT) decides where the X-ray source is placed before a scan is acquired. Robot-based and multi-axis systems reach almost any pose on a two-dimensional manifold, so a trajectory is given by the poses themselves rather than by a curve through them~\cite{Herl2020Scanning,Kingston2018}. The problem becomes acute under a sparse view budget, in industrial non-destructive evaluation with long acquisition times and in interventional imaging with strict dose and acquisition-speed constraints~\cite{Stayman2019Task,Hatamikia2022Source}. Reconstruction from few projections has been studied extensively, with analytical, iterative, and learning-based approaches~\cite{Feldkamp1984,Natterer2002Mathematics,Sidky2008Image,Jin2017Deep,Adler2018Learned,Ye2025DRACO}. Optimal source placement remains much less settled.

A natural geometric anchor is Tuy completeness~\cite{Tuy1983Inversion}, which guarantees exact cone-beam reconstruction if every plane through the region of interest intersects the source path. Discrete coverage scores use this condition by sampling plane normals on the sphere~\cite{Maier2015Discrete,Liu2012Completeness}, later augmented with attenuation-based validity constraints~\cite{Herl2020Scanning,Herl2022X}. Our prior work sharpens the resulting discrete selection problem with a soft coverage score, termed soft-Tuy in the following, together with complexity results and certified small-instance optima~\cite{Schneider2026}. Whereas coverage scores reason purely geometrically, the View Covariance Loss Swap Search (VCLS) of Lin et al.~\cite{Lin2025Tomographic} selects views by minimising a closed-form surrogate of reconstruction error and performs strongly on simulated and measured cone-beam data. Both select from a fixed discrete candidate pool, the soft-Tuy approach from arbitrary but predefined geometries, VCLS in its published implementation from a circular candidate set, although the VCL criterion itself admits arbitrary view parameters, and it additionally requires an object-specific quadratic precompute and does not model photon starvation.

A fixed pool limits the geometry to the positions it already contains, refining it only enlarges a selection problem that is already NP-complete~\cite{Schneider2026}, and every new acquisition manifold needs a new pool. A pool-free search alone does not resolve this. Linde et al.~\cite{Linde2025Automated} optimise three global orbit parameters by derivative-free differential evolution, so all views move together and none can be steered towards a direction that only a single defect requires. Making each view free instead yields $3k$ unknowns, the sample count of a population-based search grows with that dimension, and every sample costs a full objective evaluation on the volume. What is missing is a search direction that costs one objective evaluation per step instead of one per sampled perturbation, and an analytic gradient delivers exactly that.

We remove both restrictions by making every source pose an individual continuous variable of a differentiable objective. This matters when views must be refined off-grid, when the acquisition manifold is richer than a circular orbit, and when planning should account for the attenuation the object imposes on each direction.

The objective combines three differentiable terms on the source positions. The soft-Tuy surrogate, a sampled centre-point plane-coverage score, promotes angular coverage. An analytic bundle term penalises, under a monochromatic Beer--Lambert transmission model, the directions along which the transmitted intensity is lowest. A continuous extension of the VCL information criterion of Lin et al.~\cite{Lin2025Tomographic} rewards views that carry reconstruction-relevant information in the noise-free case. Coverage sees only geometry, the bundle term sees only the attenuation along its rays, and the information term assumes noise-free data. One shared objective puts all three into the same gradient step.

All results in this paper start the optimiser from the greedy soft-Tuy selector of~\cite{Schneider2026}, a geometric initialisation that needs no precompute, and refine the source poses on an idealised differentiable kinematic manifold. We realise this with four standard models, namely the single-axis circle, the limited C-arm, the two-axis gantry, and the free sphere.

This paper makes four contributions.
\begin{itemize}
\item A gradient-based framework for trajectory optimisation. Every source pose is a continuous variable on an idealised kinematic manifold, so the refinement stage needs no candidate pool.
\item A differentiable objective that puts geometric coverage, attenuation, and reconstruction information into one gradient step. The absorption term has an analytic gradient and needs no step-size calibration.
\item A matched comparison against the discrete state of the art. On the free sphere our selector reaches its image quality without any object-specific precompute, so it selects in seconds where the discrete pipeline needs minutes and its margin grows with the reconstruction grid.
\item A measured robot-based study. Pose sets planned from a prescan are acquired on a custom bench and reconstructed in a common reference frame.
\end{itemize}

All proofs, the algorithm listing, and extended tables and analyses appear in the supplementary material. %
The code is available on \href{https://github.com/Linda-SophieSchneider/Gradient-Based-Trajectory-Optimisation-over-Continuous-Poses-for-Sparse-View-Cone-Beam-CT}{GitHub}.

\section{Related Work}
\label{sec:related}
Sparse-view CT reconstruction spans analytical Feldkamp--Davis--Kress (FDK) and iterative
regularisation~\cite{Feldkamp1984,Sidky2008Image}, learned and known-operator
methods~\cite{Jin2017Deep,Han2016,Adler2018Learned,Ye2025DRACO}, and differentiable CT
operators as reusable building blocks~\cite{SunDiffCT2026}. These methods ask
how to reconstruct incomplete data; we address the complementary question of
which projections should be acquired.

Adaptive acquisition design is well established, and almost all of it selects
from a discrete pool. Batenburg et al.~\cite{Batenburg2013Dynamic}, Haque et
al.~\cite{Haque2013Adaptive}, and Lin et al.~\cite{Lin2024Edge} select angles
from information, spectral, or CAD-edge criteria; Burger et
al.~\cite{Burger2021Sequential} formulate sequential X-ray view selection as
Bayesian A- or D-optimal design. Robot and C-arm studies optimise
object-specific trajectories and out-of-plane
adjustments~\cite{Fischer2016Object,Bauer2022,Herl2021Task,Zaech2019,Thies2020}, characterise non-circular orbits~\cite{Wu2020}, evaluate families of
spherical trajectories under a fixed pose budget~\cite{Bauer2021}, and plan orbits scored on a prior reconstruction under
strong kinematic constraints~\cite{Hatamikia2020Optimization,Hatamikia2021Toward}. Learned
selectors choose views sequentially, through reinforcement-learning
policies~\cite{Shen2020,Wang2024} or a recurrent network~\cite{Yuan2024ApplicationOG};
they amortise selection over a task distribution but have so far required
training per application and acquisition manifold.

Continuous and differentiable geometry optimisation has also been studied before.
Stayman et al.~\cite{Stayman2019Task} use the
population-based, derivative-free covariance matrix adaptation evolution
strategy (CMA-ES) of Hansen and Ostermeier~\cite{Hansen2001} to maximise a task-driven
detectability index over continuous trajectory parameters
expressed as periodic or B-spline coefficients; Gang et
al.~\cite{Gang2015} optimise tube-current and reconstruction-kernel
modulation for a fixed orbit, sweeping orbital tilt over a small discrete set. Thies et
al.~\cite{Thies2023Optimizing,Thies2023Geometry} study gradient-based
geometry correction, and Fathi and van Leeuwen~\cite{Fathi2024Source}
optimise a fixed-budget laminography design. Our own differentiable-ranking
work~\cite{Schneider2023Rank} makes discrete projection selection
differentiable. However, its search variables are inclusion scores over a
fixed candidate pool, and its task-based objective must be learned per
application. Linde et al.~\cite{Linde2025Automated} search robot-CT trajectories by
differential evolution
over three continuous orbit parameters on a Radon-space completeness
criterion from a prior object model; companion work addresses space-restricted environments and few-view
budgets~\cite{Linde2024,Linde2025Few}. In closely related work first presented in mid-2025, Yoo et
al.~\cite{Yoo2026} combine a sampled Tuy-completeness score with hard
rejection of metal-artefact-inducing directions on a robotic CBCT system,
again selecting from discrete candidate positions. In cone-beam CT, however,
no prior method optimises the source poses themselves as continuous variables
of a gradient-based, fixed-budget selection.
Even where the search space is continuous, Stayman
et al. \cite{Stayman2019Task} and Linde et al. \cite{Linde2025Automated} optimise a handful of global orbit-shape parameters through a finite population of candidate solutions, whereas our optimiser updates one point per source through a smooth gradient and never instantiates any candidate set.

The closest reconstruction-driven baseline is the View Covariance Loss (VCL)
of Lin et al.~\cite{Lin2025Tomographic}. Its VCLS swap search minimises a closed-form
filtered-backprojection error surrogate over a fixed candidate set, and voxel
subsampling makes that search tractable. Discrete coverage work supplies a
complementary geometric baseline, grounding selection in Tuy's completeness
condition~\cite{Tuy1983Inversion} through sampled completeness criteria and greedy
selection~\cite{Maier2015Discrete,Liu2012Completeness,Herl2020Scanning,Herl2022X,Schneider2024IntegerOO,Schneider2026}.
In particular, our prior work~\cite{Schneider2026} shows that the decision
version of the discrete problem is NP-complete and gives a near-optimal
submodular greedy method. Clackdoyle and Noo~\cite{Clackdoyle2020Quantification} quantify cone-beam
incompleteness on a theoretically grounded graded scale, but as an evaluation
measure that is not differentiable in the source positions. We retain both
the reconstruction-driven and the geometric perspective, and make the source
positions themselves differentiable variables. Reconstruction information, soft-Tuy coverage, and attenuation are combined in one objective, without rebuilding candidate pools or object-specific precomputations for new acquisition manifolds.

\section{Methods}
\label{sec:methods}
\begin{figure*}[t]
\centering
\begin{tikzpicture}[font=\small]

\node[inner sep=0pt] (panelA) at (0,0)
  {\includegraphics[width=3.2cm]{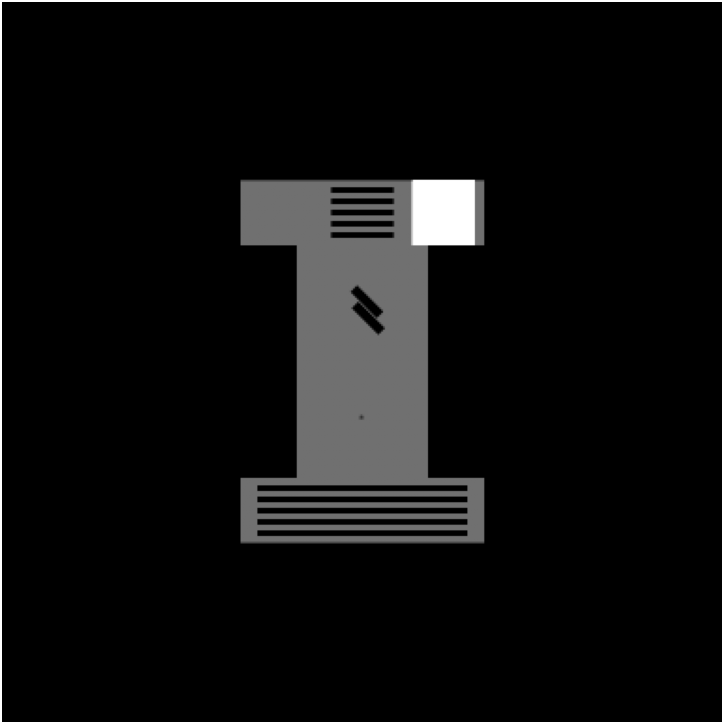}};
\node[below=0.12cm of panelA, align=center, text width=3cm]
  {\small\textbf{(a)} Attenuation model $\boldsymbol{\mu}$};

\begin{scope}[shift={($(panelA.east)+(2.6cm,0)$)}]
  \shade[inner color=white, outer color=gray!25] (0,0) circle [radius=1.5cm];
  \draw[gray!60] (0,0) circle [radius=1.5cm];
  \draw[gray!50, dashed, thin] (0,0) ellipse [x radius=1.5cm, y radius=0.42cm];
  \draw[gray!50, dashed, thin] (0,1.5) arc [start angle=90, end angle=-90,
        x radius=0.42cm, y radius=1.5cm];
  \foreach \p in {(1.5,0.3),(0.8,1.3),(-0.5,1.4),(-1.4,0.4),
                  (-1.3,-0.6),(-0.3,-1.4),(0.9,-1.1),(1.4,-0.2)}{
    \draw[->, covblue!60, thin] (0,0) -- \p;
  }
  \foreach \p in {(1.3,0.7),(0.3,1.45),(-0.9,1.1),(-1.45,0.2),
                  (-1.2,-0.9),(0.1,-1.5),(1.1,-0.9),(1.5,0.1)}{
    \fill[srcred!85] \p circle [radius=2.5pt];
  }
  \draw[->, srcred, line width=1.2pt] (1.3,0.7) -- ++(-0.21,0.40);
  \draw[->, srcred, line width=1.2pt] (0.3,1.45) -- ++(-0.44,0.09);
  \draw[->, srcred, line width=1.2pt] (-0.9,1.1) -- ++(-0.35,-0.28);
  \node[inner sep=0pt, fit={(-1.5,-1.5) (1.5,1.5)}] (panelB) {};
\end{scope}
\node[below=0.12cm of panelB, align=center, text width=3.4cm]
  {\small\textbf{(b)} Sources $\{s_i\}$ $\cdot$ Radon normals $\{n_j\}$};

\begin{scope}[shift={($(panelB.east)+(3.5cm,0)$)}]
  \tikzset{
    objbox/.style={rounded corners=3pt, draw=#1, fill=#1!12,
                   minimum width=5.3cm, minimum height=1.2cm,
                   inner sep=2pt, anchor=center}
  }
  \node[objbox=covblue] (box1) at (0,1.35) {};
  \begin{scope}[shift={($(box1.west)+(0.55,0)$)}]
    \draw[covblue!60, thin] (0,0) circle [radius=0.22cm];
    \draw[covblue!50, dashed, thin] (0,0) ellipse [x radius=0.22cm, y radius=0.07cm];
    \draw[->, covblue!70, thin] (0,0) -- (0.28,0.12);
    \draw[->, covblue!70, thin] (0,0) -- (-0.26,0.10);
    \draw[->, covblue!70, thin] (0,0) -- (0.02,0.30);
  \end{scope}
  \node[anchor=west, align=left] at ($(box1.west)+(1.15,0)$)
    {\small\bfseries\textcolor{covblue}{$\tilde{C}_{\mathrm{geo}}(V)$}\\[1pt]
     \scriptsize Gaussian-scored Tuy coverage};
  \node[objbox=absorborange] (box2) at (0,0) {};
  \begin{scope}[shift={($(box2.west)+(0.55,0)$)}]
    \fill[absorborange!30] (0,-0.05) ellipse [x radius=0.24cm, y radius=0.15cm];
    \draw[absorborange!60, thin] (0,-0.05) ellipse [x radius=0.24cm, y radius=0.15cm];
    \foreach \dx in {-0.22,-0.08,0.08,0.22}{
      \draw[absorborange!60, thin] (0,0.34) -- (\dx,-0.10);
    }
  \end{scope}
  \node[anchor=west, align=left] at ($(box2.west)+(1.15,0)$)
    {\small\bfseries\textcolor{absorborange}{$\mathcal{L}_{\mathrm{bundle}}(V)$}\\[1pt]
     \scriptsize Bundle attenuation penalty $\bar{\tau}_i$};
  \node[objbox=infogreen] (box3) at (0,-1.35) {};
  \begin{scope}[shift={($(box3.west)+(0.42,-0.22)$)}]
    \foreach \r [count=\ri from 0] in {20,50,80}{
      \foreach \c [count=\ci from 0] in {30,60,80}{
        \pgfmathtruncatemacro{\shade}{(\r+\c)/2}
        \fill[infogreen!\shade] (\ci*0.16,\ri*0.16)
              rectangle ++(0.15,0.15);
      }
    }
  \end{scope}
  \node[anchor=west, align=left] at ($(box3.west)+(1.15,0)$)
    {\small\bfseries\textcolor{infogreen}{$I_{\mathrm{vcl}}(V)$}\\[1pt]
     \scriptsize Continuous VCL information score};
  \node[below=0.18cm of box3, anchor=north]
    {\footnotesize $\mathcal{L} = \lambda_{\mathrm{cov}}\tilde{C}_{\mathrm{geo}}
      + \lambda_{\mathrm{info}}I_{\mathrm{vcl}}
      + \lambda_{\mathrm{bundle}}\mathcal{L}_{\mathrm{bundle}}$};
  \node[inner sep=0pt, fit={(box1) (box2) (box3)}] (panelC) {};
\end{scope}
\node[below=0.62cm of panelC, align=center, text width=5.6cm]
  {\small\textbf{(c)} Differentiable objective $\mathcal{L}$};

\begin{scope}[shift={($(panelC.east)+(2.6cm,0)$)}]
  \node[inner sep=0pt] (panelDtop) at (0,1.0)
    {\includegraphics[width=3.5cm, trim=232 30 20 45, clip]
      {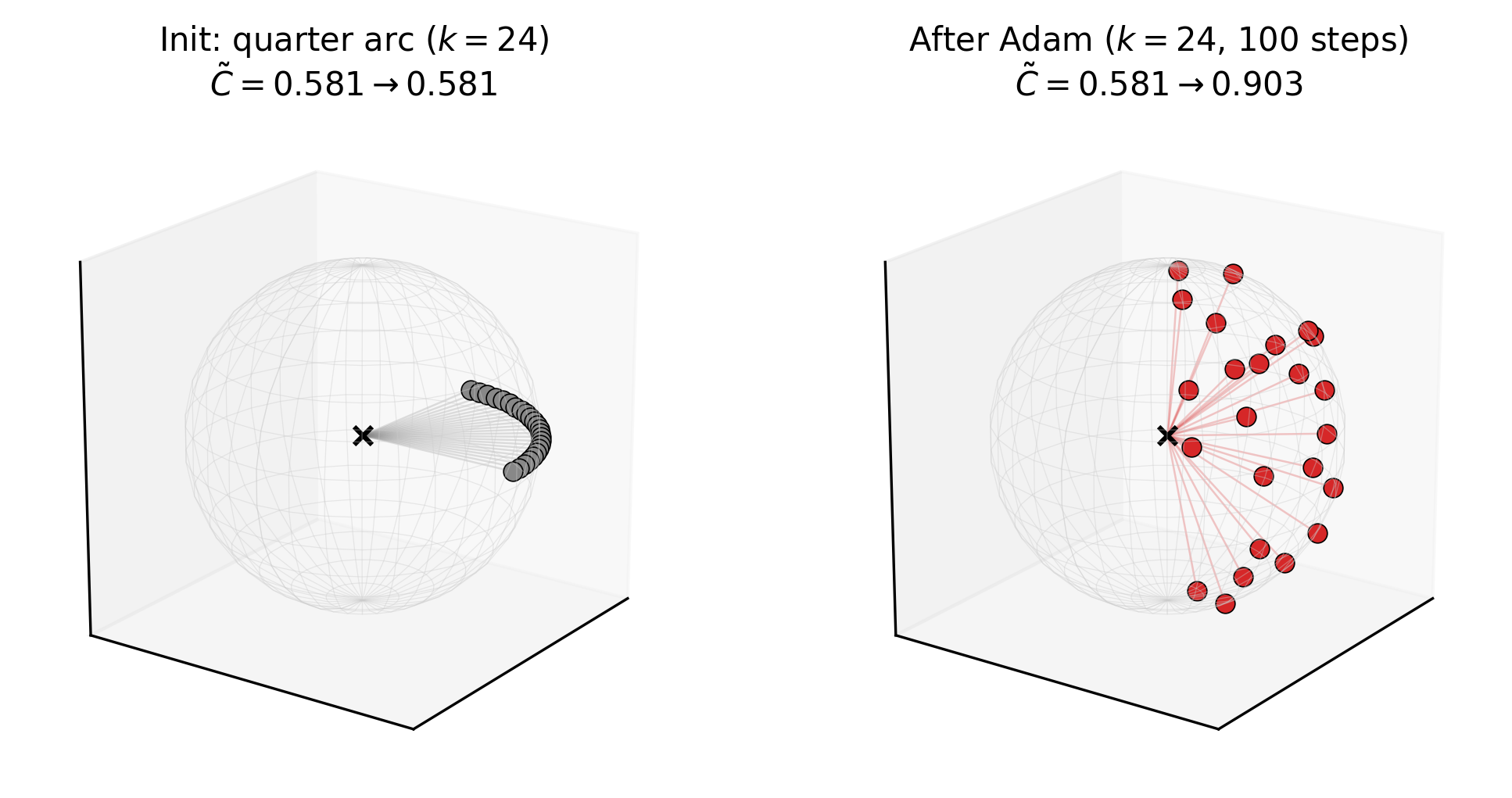}};
  \node[inner sep=0pt, below=0.15cm of panelDtop] (panelDbot)
    {\includegraphics[width=3.5cm]{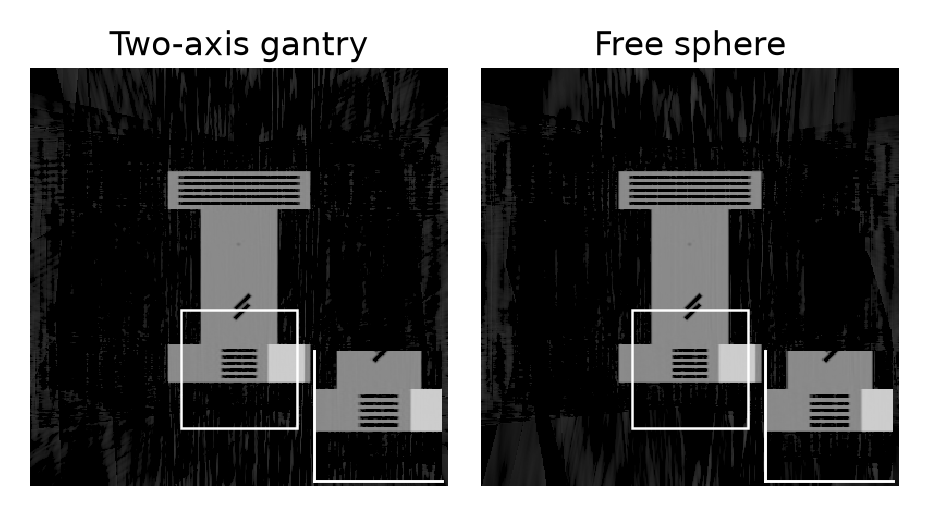}};
  \node[inner sep=0pt, fit={(panelDtop) (panelDbot)}] (panelD) {};
\end{scope}
\node[below=0.12cm of panelD, align=center, text width=3.8cm]
  {\small\textbf{(d)} Optimised $V^\star$ $\cdot$ SART reconstruction};

\draw[-{Stealth[length=5pt]}, thick, gray!70] (panelA.east) -- (panelB.west);
\draw[-{Stealth[length=5pt]}, thick, gray!70] (panelB.east) -- (panelC.west);
\draw[-{Stealth[length=5pt]}, thick, gray!70] (panelC.east) -- (panelD.west);
\draw[-{Stealth[length=4pt]}, gray!60]
  (panelC.north) to[out=90, in=90, looseness=0.6]
  node[midway, above, font=\scriptsize, text=gray!70]
  {Adam $\partial\mathcal{L}/\partial\theta_i$} (panelB.north);

\end{tikzpicture}
\caption{Overview of the gradient-based trajectory-optimisation pipeline.
  \textbf{(a)}~Attenuation model $\boldsymbol{\mu}$.
  \textbf{(b)}~The viewing sphere carries two families of symbols. Red dots are the source positions $s_i$, and blue arrows from the centre are the sampled Radon normals $n_j$, one for each plane whose coverage is scored. Thick red arrows are the coverage gradient, which is tangential to the sphere by Proposition~\ref{prop:tangential_grad}. Every source carries such a gradient, and three are drawn so that the panel stays readable.
  \textbf{(c)}~The objective combines the soft-Tuy coverage surrogate, the attenuation-aware bundle penalty, and the continuous VCL information score; Adam updates the kinematic parameters on the manifold.
  \textbf{(d)}~Two outputs from separate runs. Above, red dots are the optimised source positions $V^\star$ of a coverage-only free-sphere run at $k{=}24$, thin red lines are their rays to the region of interest, and the black cross marks its centre. Below, SART reconstructions at $k{=}80$ for the two-axis gantry and the free sphere, named in the image.}
\label{fig:method_overview}
\end{figure*}

Selecting X-ray source positions for cone-beam CT is usually posed as a combinatorial search over a finite candidate pool.
We instead treat the source positions as continuous variables on a kinematic manifold and express coverage, absorption, and view information as differentiable functions of these positions. View selection then becomes a gradient-based optimisation problem, as illustrated in \cref{fig:method_overview}.
\Cref{sec:methods_formulation} introduces the differentiable kinematic parametrisation. The objective combines a smooth soft-Tuy coverage surrogate with a closed-form tangential gradient in \crefrange{sec:methods_coverage}{sec:methods_gradient}, an analytic bundle absorption term in \cref{sec:methods_path}, and a continuous extension of the VCL score of Lin et al.\ in \cref{sec:methods_vcl}.
The combined objective is maximised with Adam under a cosine schedule and manifold-aware refinement, as detailed in \cref{sec:methods_combined}; the complete algorithm is provided in the supplementary material.

\subsection{Problem Formulation}
\label{sec:methods_formulation}
We build on the discrete framework of~\cite{Schneider2026}, which maximises a soft Tuy-coverage surrogate over a finite candidate set. Our change is the search space, optimising the source positions themselves.
Let $V = \{s_1, \ldots, s_k\}$ with $s_i \in \mathbb{R}^3$ be the current view set. Further, let $x_{\mathrm{roi}} \in \mathbb{R}^3$ denote the region-of-interest (ROI) centre, $r_{\mathrm{sid}}>0$ the source-to-isocentre distance, and $\{n_j\}_{j=1}^{z}$ with $n_j \in \mathbb{S}^2$ a Fibonacci lattice of Radon-plane normals on the unit sphere~\cite{Herl2020Scanning}, with $z$ samples in total. The $n_j$ are the normals of the Radon planes through the ROI centre; since all these planes share that anchor, only the orientation remains free. Each pose is described by a kinematic parameter vector $\theta_i \in \mathbb{R}^{p}$, where $p$ depends on the chosen acquisition manifold. Source positions are parametrised through a differentiable kinematic map
\begin{equation}
  s_i = \Gamma(\theta_i),
\label{eq:kinematics}
\end{equation}
where $\Gamma$ encodes scanner kinematics. The manifolds used here share the map
\begin{equation}
  \Gamma_{\mathrm{2ax}}(\theta,\varphi)
  = r_{\mathrm{sid}}
  \begin{bmatrix}
    -\cos\varphi\sin\theta &
     \cos\varphi\cos\theta &
     \sin\varphi
  \end{bmatrix}^{\!\top}.
\label{eq:kinematics_two_axis}
\end{equation}
The single-axis circle is $\Gamma_{\mathrm{circ}}(\theta)=
\Gamma_{\mathrm{2ax}}(\theta,0)$ with $\theta$ periodic modulo $2\pi$.
The unconstrained two-axis gantry uses the same periodic map; the canonical
reporting ranges are $\theta\in[-\pi,\pi)$ and
$\varphi\in[-\pi/2,\pi/2]$.
An envelope-restricted axis is expressed by a chart that represents the
envelope itself rather than by restricting an unbounded one. For an axis
admitted on $[a,b]$ we optimise a free coordinate $\tilde\alpha\in\R$ and
read the angle off
\[
  \alpha(\tilde\alpha)
  = \tfrac{1}{2}(a+b) + \tfrac{1}{2}(b-a)\tanh\tilde\alpha ,
\]
so every iterate is admissible by construction, the gradient is everywhere
tangent to the reachable set, and no feasibility repair enters the update. The
limited C-arm uses this chart on both axes, with
$\theta\in[-110^\circ,110^\circ]$ and $\varphi\in[-45^\circ,45^\circ]$;
the elevation band of our measured bench uses it on $\varphi$ alone and keeps
$\theta$ periodic, since a bounded chart would break the wrap-around of a full
revolution. Throughout, the isocentre coincides with the ROI centre $x_{\mathrm{roi}}$, which the two-axis map places at the origin. The free sphere is
$\{s \mid \|s-x_{\mathrm{roi}}\|=r_{\mathrm{sid}}\}$; its ambient-coordinate
update is retracted by
$\Pi_{\mathbb S^2}(s)=x_{\mathrm{roi}}+
r_{\mathrm{sid}}(s-x_{\mathrm{roi}})/\|s-x_{\mathrm{roi}}\|$.
Each source pose induces the standard cone-beam detector geometry, an untilted flat detector at fixed source-to-detector distance, centred on the principal ray through $x_{\mathrm{roi}}$ and oriented by $u_i \propto e_z \times (s_i - x_{\mathrm{roi}})$, with $u_i=e_x$ at the two poles where the cross product vanishes. These detector parameters remain fixed, so the optimisation acts only on the source positions. The continuous view-selection problem is then
\begin{equation}
  \max_{\theta_1, \ldots, \theta_k}\;
  \mathcal{L}\!\left(\Gamma(\theta_1), \ldots, \Gamma(\theta_k)\right),
\label{eq:problem}
\end{equation}
with the objective $\mathcal{L}$ built term by term in the following subsections.


\subsection{Differentiable Soft Coverage}
\label{sec:methods_coverage}
For source $s_i \in V$ and ROI centre $x_{\mathrm{roi}}$, define the displacement, distance, and unit viewing direction by
\begin{equation}
  r_i = s_i - x_{\mathrm{roi}}, \qquad
  \rho_i = \norm{r_i}, \qquad
  d_i = r_i/\rho_i.
\label{eq:direction_decomp}
\end{equation}
Here, $d_i$ is the unit viewing direction from the ROI centre to source $s_i$. For each Radon-plane normal $n_j$, the quantity $g_{ij} = n_j^\top d_i$ measures how close this viewing direction is to the corresponding plane; exact alignment gives $g_{ij} = 0$. The hinge-based coverage score of~\cite{Schneider2026} is piecewise linear, not differentiable at the tolerance boundary, and carries no gradient information beyond it. Therefore, we replace it by the smooth Gaussian kernel
\begin{equation}
  \psi_\sigma(g) = \exp\!\left(-g^2 / 2\sigma^2\right),
\label{eq:gaussian_score}
\end{equation}
with Gaussian bandwidth $\sigma$. We match it to a target tolerance $\tau = \sin(\Delta\gamma)$ by imposing $\psi_\sigma(\tau) = \epsilon$, which yields
\begin{equation}
  \sigma = \tau / \sqrt{2\log(1/\epsilon)}.
\label{eq:sigma_match}
\end{equation}
Here, $\Delta\gamma\in(0,\pi/2)$ sets the angular tolerance and $\epsilon\in(0,1)$ its corresponding score, so that $\sigma>0$ is well defined. We then accumulate the contributions of all views in $V$ along direction $n_j$ as $\Sigma_j(V) = \sum_{i=1}^{k} \psi_\sigma(g_{ij})$. Applying the smooth saturation function $\phi(x) = 1 - e^{-x}$ gives the geometry-only coverage surrogate
\begin{equation}
  \tilde{C}_{\mathrm{geo}}(V)
  = \frac{1}{z} \sum_{j=1}^{z} \phi\!\left(\Sigma_j(V)\right)
  \in [0, 1).
\label{eq:geo_obj}
\end{equation}
$\tilde{C}_{\mathrm{geo}}$ averages the saturated coverage over the $z$ sampled Radon-plane normals and is the smooth analogue of the hinge-based coverage of~\cite{Schneider2026}. Because every view contributes non-negatively to each $\Sigma_j$ and the saturation $\phi$ is increasing and concave, $\tilde{C}_{\mathrm{geo}}$ is monotone and submodular as a set function in the selected sources~\cite{Nemhauser1978analysis}.
\subsection{Closed-Form Tangential Coverage Gradient}
\label{sec:methods_gradient}
The coverage term depends on the viewing direction $d_i$, not directly on the source-to-ROI distance $\rho_i$. The first step is therefore to understand how $d_i$ changes when the source position moves. Only tangential motion changes the viewing direction; radial motion changes the distance to the ROI but not the direction itself. This geometric separation is captured by the Jacobian of the unit direction with respect to the source position.
\begin{proposition}[Tangential Jacobian]
\label{prop:jacobian}
The derivative of the unit viewing direction with respect to the source position is the orthogonal projection onto the tangent plane at $d_i$, scaled by the inverse source-to-ROI distance, and is given by
\[
\partial d_i/\partial s_i = \rho_i^{-1}(I_3 - d_i d_i^\top) = \rho_i^{-1} P_{d_i}^{\perp},
\]
where $P_{d_i}^{\perp} = I_3 - d_i d_i^\top$ is the orthogonal projector onto the tangent plane of $\mathbb{S}^2$ at $d_i$.
\end{proposition}
The proof is provided in the supplementary material.

Applying this Jacobian to $g_{ij} = n_j^\top d_i$ gives the gradient of the Radon-plane mismatch with respect to the source position,
\begin{equation}
  \nabla_{s_i} g_{ij}
  = \frac{\partial d_i}{\partial s_i}^{\!\top}\!n_j
  = \frac{1}{\rho_i}\bigl(n_j - g_{ij}\,d_i\bigr),
\label{eq:grad_g}
\end{equation}
where the second equality uses $P_{d_i}^{\perp}n_j = n_j - g_{ij} d_i$. Differentiating the Gaussian score $\psi_\sigma$ then gives the single-direction coverage gradient
\begin{equation}
  \nabla_{s_i}\psi_\sigma(g_{ij})
  = -\frac{g_{ij}}{\sigma^2}\,\psi_\sigma(g_{ij})\,
    \frac{1}{\rho_i}\bigl(n_j - g_{ij}\,d_i\bigr).
\label{eq:grad_psi}
\end{equation}
The kernel derivative
$\lvert\psi_\sigma'(g)\rvert=|g|\psi_\sigma(g)/\sigma^2$
vanishes at perfect coverage and peaks at $|g|=\sigma$.
Applying $\phi'(\Sigma_j)=e^{-\Sigma_j}$ yields the full coverage gradient
\begin{align}
  \nabla_{s_i} \tilde{C}_{\mathrm{geo}}
  &= \frac{1}{z} \sum_{j=1}^{z}
       e^{-\Sigma_j}\,
       \nabla_{s_i}\psi_\sigma(g_{ij}) \nonumber\\
  &= -\frac{1}{z\rho_i} \sum_{j=1}^{z}
       \frac{g_{ij}\,\psi_\sigma(g_{ij})\,e^{-\Sigma_j}}{\sigma^2}\,
       \left(n_j - g_{ij}\,d_i\right).
\label{eq:grad_C_full}
\end{align}
The saturation factor $e^{-\Sigma_j}$ discourages further investment in well-covered directions, the scalar factor $-(g_{ij}/\sigma^2)\psi_\sigma(g_{ij})$ weights each direction by the kernel sensitivity, and the direction factor $n_j - g_{ij}\,d_i = P_{d_i}^{\perp} n_j$ lies in the tangent plane at $d_i$, so the full coverage gradient remains tangential to the viewing sphere.
\begin{proposition}[Tangential Coverage Gradient]
\label{prop:tangential_grad}
The coverage gradient with respect to the source position is orthogonal to the viewing direction, so that
\[
\nabla_{s_i} \tilde{C}_{\mathrm{geo}} \perp d_i
\]
for every source position $s_i$ with $\rho_i > 0$.
\end{proposition}
The proof is provided in the supplementary material.

For a kinematic parametrisation $s_i = \Gamma(\theta_i)$, the parameter-space gradient is
\begin{equation}
  \nabla_{\theta_i} \tilde{C}_{\mathrm{geo}}
  = \bigl(\partial\Gamma(\theta_i)/\partial\theta_i\bigr)^{\!\top}
    \nabla_{s_i}\tilde{C}_{\mathrm{geo}},
\label{eq:kinematic_chain}
\end{equation}
which maps the source-space gradient to the kinematic coordinates by the chain rule.

The centred surrogate of~\cref{eq:geo_obj} is the form used in all virtual studies. The measured study of \cref{sec:results_real} averages the same construction over sampled ROI points instead of a single centre; at the $5$~mm ROI size and $\Delta\gamma = 15^{\circ}$ smoothing used here, the two forms agree in attained coverage to within $10^{-3}$.
\subsection{Analytic Bundle Absorption Penalty}
\label{sec:methods_path}
Geometry-only coverage treats every direction as equally informative, although long paths through dense material transmit fewer photons even where coverage is good. We therefore add an object-aware penalty, a forward projection along a small bundle of rays from each source to a target patch around the ROI. Unlike a general projection operator~\cite{SunDiffCT2026}, which is differentiated with respect to the volume, the rays are parametrised directly in $s_i$, so the source gradient is available in closed form at the cost of a few tens of rays per view.

To place a square patch of bundle targets around the ROI, we construct for source $s_i$ an orthonormal frame $(\hat u_i, \hat v_i, d_i)$ with $(\hat u_i, \hat v_i)$ spanning the tangent plane $d_i^\perp$, using a cross-product construction with a reference axis $\hat r_i$,
\begin{equation}
  \hat u_i = \frac{d_i \times \hat r_i}{\norm{d_i \times \hat r_i}},
  \qquad
  \hat v_i = d_i \times \hat u_i,
\label{eq:bundle_frame}
\end{equation}
where $\hat r_i = \hat z$ if $|d_i^\top \hat z| < 0.95$ and $\hat r_i = \hat y$ otherwise; the fallback avoids the polar degeneracy of the frame construction. Let $\{(g_u^{(r)}, g_v^{(r)})\}_{r=1}^{|\mathcal R|}$ be a uniform $n_u \times n_v$ grid on $[-1,1]^2$, and let $R_{\mathrm{roi}}$ denote the half-width of the square bundle target patch in the plane $d_i^\perp$. The corresponding target points are
\begin{equation}
  c_i^{(r)} = x_{\mathrm{roi}} + R_{\mathrm{roi}}\,
    \left(g_u^{(r)}\hat u_i + g_v^{(r)}\hat v_i\right),
  \quad r = 1, \ldots, |\mathcal R|,
\label{eq:bundle_targets}
\end{equation}
and the line integral along ray $r$, parametrised by $p_i^{(r)}(t) = s_i + t\,(c_i^{(r)} - s_i)$ with $t \in (0, 1]$, is
\begin{equation}
  \tau_i^{(r)}
  = \norm{c_i^{(r)} - s_i} \int_{0}^{1}
      \mu\!\left(p_i^{(r)}(t)\right)\,\mathrm{d}t,
\label{eq:bundle_ray}
\end{equation}
where $\mu$ is the attenuation field of the reference volume available at planning time. The bundle-mean line integral
\begin{equation}
  \bar\tau_i = \frac{1}{|\mathcal R|}
                 \sum_{r=1}^{|\mathcal R|}\tau_i^{(r)}
\label{eq:bundle_mean}
\end{equation}
yields one dimensionless optical-depth score per source.
\Cref{eq:bundle_ray} is evaluated with the midpoint rule on the
sub-segment $[t_{\mathrm{in}}^{(r)}, t_{\mathrm{out}}^{(r)}] \subseteq [0,1]$
of each ray inside the reference volume, obtained by intersecting the ray
with the volume bounding box, at the $N_t$ regularly spaced samples
$t_n = t_{\mathrm{in}}^{(r)} + (n-\tfrac{1}{2})\,
(t_{\mathrm{out}}^{(r)}-t_{\mathrm{in}}^{(r)})/N_t$. The attenuation values
are obtained by trilinear interpolation of $\mu$. Interpolation is what makes
the quadrature differentiable with respect to the source position, since the
sample points move with the source and a nearest-neighbour lookup would be
piecewise constant with vanishing gradient almost everywhere. The resulting
approximation is
\begin{equation}
  \tau_i^{(r)} \approx \frac{\norm{c_i^{(r)} - s_i}\,
  \bigl(t_{\mathrm{out}}^{(r)}-t_{\mathrm{in}}^{(r)}\bigr)}{N_t}
  \sum_{n=1}^{N_t} \mu\!\left(p_i^{(r)}(t_n)\right).
\end{equation}
Restricting the quadrature to the ray--volume intersection loses nothing
physically, because $\mu$ vanishes outside the reference volume, and it
ensures every sample carries attenuation information rather than reading
clamped border values. A pre-registered convergence study against a $4096$-sample reference, reported in the supplementary material, fixes $N_t=256$; at this value the integrals, their source gradients, and the induced view rankings are converged.
For this trilinear interpolation, let $\bm q(p) = (p - p_0)/\Delta_v$ denote the volume-grid coordinates of a world point $p$, where $p_0$ is the volume origin and $\Delta_v$ the voxel size. We then split $\bm q$ into the floor index $\lfloor\bm q\rfloor \in \mathbb{Z}^3$ and the fractional offsets $\bm f(p) = \bm q - \lfloor\bm q\rfloor \in [0,1)^3$. The floor index identifies the voxel cell that contains $p$, and the fractional offsets locate $p$ inside that cell. The interpolated value
\begin{equation}
\begin{aligned}
    \mu(p) &= \sum_{c \in \{0,1\}^3} \mu(v_c)\,w_c(p), \\
    w_c(p) &= \prod_{\xi}\!\bigl(c_\xi f_\xi + (1\!-\!c_\xi)(1\!-\!f_\xi)\bigr),
\end{aligned}
\end{equation}
is differentiable within each voxel cell with respect to the fractional offsets $\bm f(p)$. Here, $v_c=\lfloor\bm q\rfloor+c$ are the eight cell corners and $\xi$ indexes the axes. Differentiating the weights $w_c(p)$ therefore gives the spatial gradient
\begin{equation}
\begin{aligned}
\frac{\partial\mu(p)}{\partial p}
=
\Delta_v^{-1}\sum_c \mu(v_c)\,\frac{\partial w_c}{\partial\bm f},
\end{aligned}
\end{equation}
where each partial derivative follows by differentiating one linear factor of $w_c$. Coordinates are clipped independently to the closed volume bounds before interpolation. This amounts to a constant boundary extension, where an outside-volume sample takes the nearest boundary value and its derivative in a clipped coordinate is zero. Within the volume, floor indices remain constant under differentiation and enter only through the gathered corner values $\mu(v_c)$.

The bundle mean $\bar\tau_i$ is differentiated end-to-end with respect to $s_i$. The hard reference-axis branch in~\cref{eq:bundle_frame} makes the construction piecewise differentiable, with a possible discontinuity at $|d_i^\top\hat z|=0.95$.

We use the raw attenuation score directly instead of wrapping it in a bounded gate, since a gate such as $\nu_i = \exp(-\alpha\bar\tau_i)$ saturates on highly absorbing directions and its gradient vanishes there as well. We define the bundle penalty as
\begin{equation}
  \mathcal{L}_{\mathrm{bundle}}(V)
  = -\,\frac{1}{k}\sum_{i=1}^{k}\bar\tau_i.
\label{eq:bundle_loss}
\end{equation}

A natural alternative is to finite-difference the source coordinates through an existing cone-beam forward projector, and Supplementary \cref{supp-sec:supp_ablation} reports that comparison in matched form with a step size of $\varepsilon=0.5$~mm selected by an explicit sweep. This mode acts on the absorption objective alone and is separate from the VCL geometry difference of \cref{sec:methods_vcl}. The analytic bundle instead provides the gradient directly through the attenuation field $\mu$, without differentiating through a full projector.
\subsection{Continuous View Covariance Loss}
\label{sec:methods_vcl}
The VCL of Lin et al.~\cite{Lin2025Tomographic} is a sample-based approximation to the normalised mean-squared error of a linear filtered-backprojection (FBP) reconstruction at the candidate view set.

Let $\bm A_s \in \mathbb{R}^{m \times N}$ be the cone-beam forward projector at source position $s$, $\bm H$ a one-dimensional ramp filter applied along each detector row, and $\bm A_s^\top$ its matched-footprint adjoint. The per-view FBP operator is
\begin{equation}
  \bm T_s = \bm A_s^\top \bm H \bm A_s.
\label{eq:vcl_operator}
\end{equation}
For a fixed voxel set $\Omega\subset\{1,\ldots,N\}$ of size $n=\lceil r_1 N\rceil$, drawn uniformly at random once per run with sampling fraction $r_1$, let $\bm S_\Omega\in\{0,1\}^{n\times N}$ be the corresponding sampling operator and $\bm x_\Omega=\bm S_\Omega\bm x$. We form the normalised per-view response
\begin{equation}
  \begin{aligned}
    \bm q_i(s_i)
    &=
    \frac{\bm S_\Omega\bm T_{s_i}\bm x}
         {\norm{\bm S_\Omega\bm T_{s_i}\bm x}+\varepsilon_{\mathrm n}},
    \\
    \bm Q(V)
    &=
    \begin{bmatrix}
      \bm q_1(s_1)^\top\\[-0.2em]
      \vdots\\[-0.2em]
      \bm q_k(s_k)^\top
    \end{bmatrix}
    \in\mathbb{R}^{k\times n},
  \end{aligned}
\label{eq:vcl_basis}
\end{equation}
with $\varepsilon_{\mathrm n}=10^{-9}$. The score requires $\norm{\bm x_\Omega}>0$, and the implementation rejects a zero-norm sample. The regularised view--view and view--reference correlations used by the optimiser are
\begin{equation}
  \begin{aligned}
    \boldsymbol R(V)
    &=
    \bm Q(V)\bm Q(V)^\top+\lambda_{\mathrm R}\bm I_k,
    &\lambda_{\mathrm R}&=10^{-3},
    \\
    \boldsymbol\gamma(V)
    &=
    \bm Q(V)\frac{\bm x_\Omega}{\norm{\bm x_\Omega}}.
  \end{aligned}
\label{eq:vcl_R_gamma}
\end{equation}
The continuous information score is
\begin{equation}
  I_{\mathrm{vcl}}(V)
  =
  \boldsymbol\gamma(V)^{\!\top}
  \boldsymbol R(V)^{-1}
  \boldsymbol\gamma(V)
  \;\in\; [0,1].
\label{eq:vcl_def}
\end{equation}
The bounds follow from $\bm Q\bm Q^\top\preceq\boldsymbol R$. In the unregularised limit, this is the squared cosine between the sampled reference and its projection onto the span of the per-view responses. Row normalisation leaves that ideal score invariant but fixes the scale of the ridge term in~\cref{eq:vcl_R_gamma}. The discrete VCLS of~\cite{Lin2025Tomographic} maximises this score over a candidate pool by swap search; here, $I_{\mathrm{vcl}}$ enters the continuous objective.

The matched-footprint backprojector exposed by \texttt{diffct}~\cite{SunDiffCT2026} provides a vector--Jacobian product (VJP) with respect to the projection data but no complete geometry VJP. Propagating only through the forward projection would omit the first term in
\begin{equation}
  \mathrm D_s\!\left(\bm A_s^\top\bm H\bm A_s\bm x\right)
  =
  \mathrm D_s(\bm A_s^\top)\bm H\bm A_s\bm x
  +
  \bm A_s^\top\bm H\,\mathrm D_s(\bm A_s)\bm x.
\label{eq:vcl_geometry_chain}
\end{equation}
We therefore estimate the geometry derivative of the complete normalised response in~\cref{eq:vcl_basis} by a central difference, which re-evaluates the entire pipeline at perturbed source positions and thus captures both terms of~\cref{eq:vcl_geometry_chain},
\begin{equation}
  \begin{aligned}
    \widehat{\frac{\partial\bm q_i}{\partial s_{i,a}}}
    &=
    \frac{\bm q_i(s_i+\varepsilon_{\mathrm{vcl}}\bm e_a)
         -\bm q_i(s_i-\varepsilon_{\mathrm{vcl}}\bm e_a)}
         {2\varepsilon_{\mathrm{vcl}}},
    \\
    &\hspace{7em} a\in\{1,2,3\},
  \end{aligned}
\label{eq:vcl_geometry_vjp}
\end{equation}
where $\varepsilon_{\mathrm{vcl}}=0.5$~mm. Each perturbation rebuilds the source-dependent detector frame and re-evaluates the complete response pipeline. Since response row $i$ depends only on source $i$, every view can be perturbed along the same coordinate simultaneously. The VJP therefore uses six additional basis evaluations per Adam step, independent of $k$.

The outer VCL derivative remains analytical, using a custom vector--Jacobian product for $f(\boldsymbol\gamma,\boldsymbol R)=\boldsymbol\gamma^\top\boldsymbol R^{-1}\boldsymbol\gamma$ instead of autograd through the explicit inverse.
\begin{proposition}[VJP of the VCL Quadratic Form]
\label{prop:vcl_vjp}
For symmetric positive-definite $\boldsymbol R$, as in~\cref{eq:vcl_R_gamma}, let
\[
f(\boldsymbol\gamma, \boldsymbol R) = \boldsymbol\gamma^\top\boldsymbol R^{-1}\boldsymbol\gamma,
\qquad
\boldsymbol u = \boldsymbol R^{-1}\boldsymbol\gamma.
\]
Then
\begin{equation}
  \frac{\partial f}{\partial \boldsymbol\gamma} = 2\boldsymbol u,
  \qquad
  \frac{\partial f}{\partial \boldsymbol R}
    = -\boldsymbol u\,\boldsymbol u^{\!\top}.
\label{eq:vcl_vjp}
\end{equation}
\end{proposition}
The proof follows from the differential identity for $\boldsymbol R^{-1}$ and is provided in the supplementary material.

The outer reverse pass therefore requires a single linear solve $\boldsymbol R\boldsymbol u = \boldsymbol\gamma$ and one outer product $\boldsymbol u\boldsymbol u^\top$, with no $k \times k$ inverse ever materialised explicitly. The resulting cotangent with respect to the rows of $\bm Q$ is contracted with~\cref{eq:vcl_geometry_vjp} to obtain the source update.
\subsection{Composite Objective and Manifold Refinement}
\label{sec:methods_combined}
The terms combine additively into the ascent objective
\begin{equation}
  \mathcal{L}(V)
  = \lambda_{\mathrm{cov}}\,\tilde{C}_{\mathrm{geo}}(V)
  + \lambda_{\mathrm{vcl}}\,I_{\mathrm{vcl}}(V)
  + \lambda_{\mathrm{bundle}}\,\mathcal{L}_{\mathrm{bundle}}(V),
\label{eq:loss_combined}
\end{equation}
with the continuous VCL score $I_{\mathrm{vcl}}$ of~\cref{eq:vcl_def}, which
carries no noise model. Attenuation-awareness, a monochromatic
photon-starvation proxy, enters through the explicit absorption penalty
$\lambda_{\mathrm{bundle}}\mathcal{L}_{\mathrm{bundle}}$.

\Cref{eq:loss_combined} covers all continuous variants. Inactive
terms are switched off by setting their weights to zero, and the weights are
fixed during a given optimisation run.

A discrete swap search can score the same weighted sum, but every candidate swap then pays for a full evaluation of each active term.

We maximise $\mathcal{L}$ by differentiating through the chosen kinematic map $s_i = \Gamma(\theta_i)$ and updating the parameter vectors $\theta_i$ with Adam~\cite{Kingma2015}, using a cosine learning-rate schedule with patience-based early stopping on the best objective seen so far; the complete loop is listed in the supplementary material. The concrete settings are given in~\cref{sec:setup_hyperparams}.

The Adam learning rate is a displacement in the units of the chart, so mixed-chart implementations advance poses at different angular rates per step. Scaling the Cartesian chart by the sphere radius removes the discrepancy, and we report all comparisons under that scaling. Feasibility on the circle and the C-arm is enforced by $\Gamma$ itself; on the free sphere each source is re-normalised to the viewing sphere after each step by the closest-point projection $\Pi_{\mathbb S^2}$.

The Adam loop can be initialised from a converged discrete VCLS solution or from a discrete soft-coverage greedy solution, and the two starts place the optimiser in different basins of the objective landscape. All reported results use the greedy start.

\section{Experimental Setup}
\label{sec:setup}
We compare continuous and discrete selection at matched view budgets on two virtual benchmarks, with and without photon noise, and ablate the derivative estimator and optimiser. Further studies test kinematic constraints, sparse-prescan planning, selection cost, and physical acquisition on a robot bench.

\begin{table*}[t]
\caption{Configuration of the three test objects. The flange is simulated under a monochromatic model, and the camera detector is mean-pooled by a factor of four before reconstruction.}
\label{tab:setup_objects}
\centering
\small
\setlength{\tabcolsep}{7pt}
\begin{tabular}{@{}l ccc@{}}
\toprule
Specification & Defrise flange & Fuel nozzle & Digital camera \\
\midrule
Material & aluminium, steel & 316L stainless steel & plastics, circuit boards, metal \\
Max.\ photon energy (keV) & monochromatic & $180$ & $175$ \\
Detector dimensions (pixels) & $256\times256$ & $1024\times1024$ & $3072\times3072$ \\
Pixel pitch in detector (mm) & $0.5$ & $0.2$ & $0.139$ \\
Source-to-detector distance (mm) & $900$ & $808.508$ & $1995.6$ \\
Source-to-isocentre distance (mm) & $500$ & $243.307$ & $997.1$ \\
Reconstruction grid & $384^{3}$ & $512^{3}$ & $768^{3}$ \\
Voxel pitch in object (mm) & $0.3$ & $0.118$ & $0.278$ \\
Reference volume & phantom ground truth & FDK of $1050$ views & FDK of $1200$ views \\
\bottomrule
\end{tabular}
\end{table*}

\subsection{Virtual Benchmarks and Test Objects}
\label{sec:setup_phantoms}
\begin{figure}[!t]
\centering
\includegraphics[width=0.99\columnwidth]{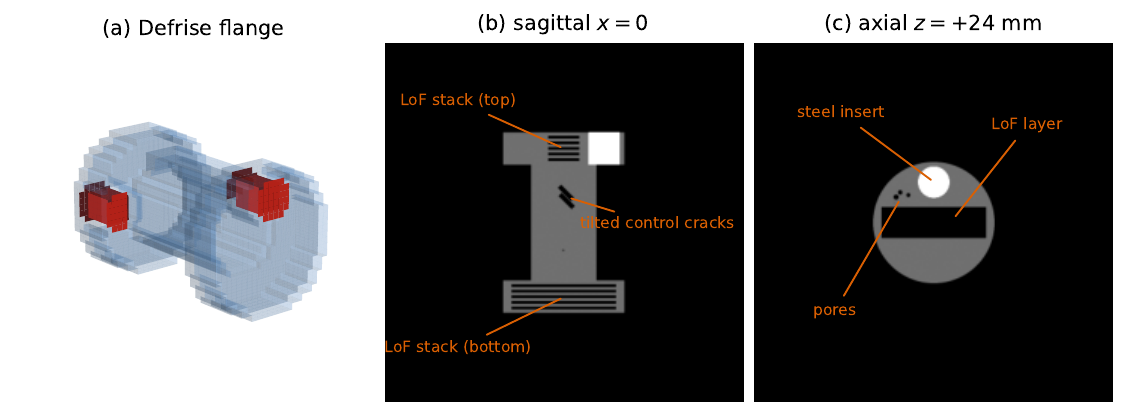}
\caption{The Defrise flange phantom. (a)~Three-dimensional view of the aluminium body in translucent blue with two press-fit steel inserts in red. (b)~Central sagittal slice with the two LoF stacks far off the source plane and the two tilted control cracks in the column. (c)~Axial slice through the upper flange disk with the LoF layer, the steel insert, and the pore cluster in the attenuation shadow of the insert.}
\label{fig:defrise_flange}
\end{figure}

\begin{figure}[!t]
\centering
\includegraphics[width=\columnwidth]{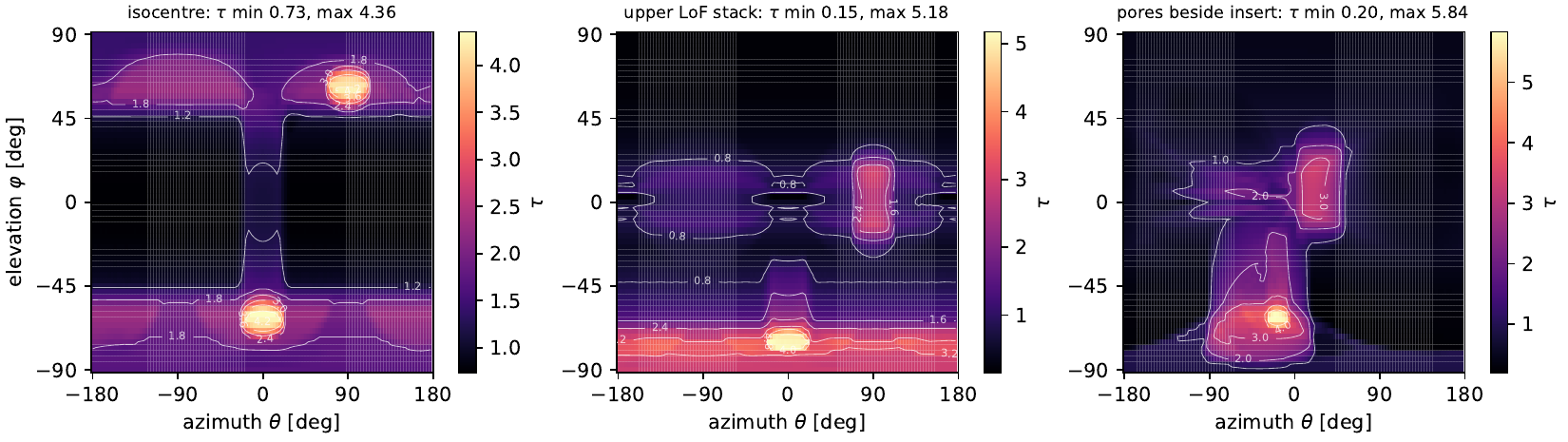}
\caption{Directional absorption of the Defrise flange, shown as the bundle path integral $\bar\tau$ towards three target points as a function of the source azimuth and elevation on the acquisition sphere. Dark is transparent. Towards the isocentre the transparent region is the equatorial band, since only the column lies in the way there. Towards an off-plane defect the minimum moves into the hemisphere of that defect and leaves a narrow equatorial notch. We aim the term at the isocentre because that choice needs no assumption about which defect matters.}
\label{fig:absorption_map}
\end{figure}

The Defrise flange (\cref{fig:defrise_flange}) tests missing elevation: planar lack-of-fusion (LoF) voids in its off-plane disks are invisible to an equatorial orbit, while tilted cracks and central pores serve as reachable controls. Steel inserts create attenuation shadows. \Cref{tab:setup_objects,fig:absorption_map} gives acquisition specifications and directional absorption; Supplementary \cref{supp-sec:supp_phantom_details} provides the full phantom design.

The Oak Ridge National Laboratory (ORNL) fuel nozzle~\cite{Ziabari2025ORNLData} is a near-axisymmetric dense-metal counterpart. Its FDK volume provides the attenuation reference; all virtual comparisons forward-project the reference volumes with matched \texttt{diffct} operators~\cite{SunDiffCT2026} and reconstruct using $15$ simultaneous algebraic reconstruction technique (SART) iterations~\cite{Andersen1984SART}, relaxation $0.9$, and non-negativity. Supplementary \cref{supp-sec:supp_virtual_protocol} gives implementation details.

The photon-noise model and its evaluation on both objects are reported with the complete results in Supplementary \cref{supp-sec:supp_noise}.

\subsection{Measured Robot-Based CT Study}
\label{sec:setup_real}
\begin{figure}[!t]
\centering
\subfloat[]{%
\includegraphics[width=0.48\columnwidth]{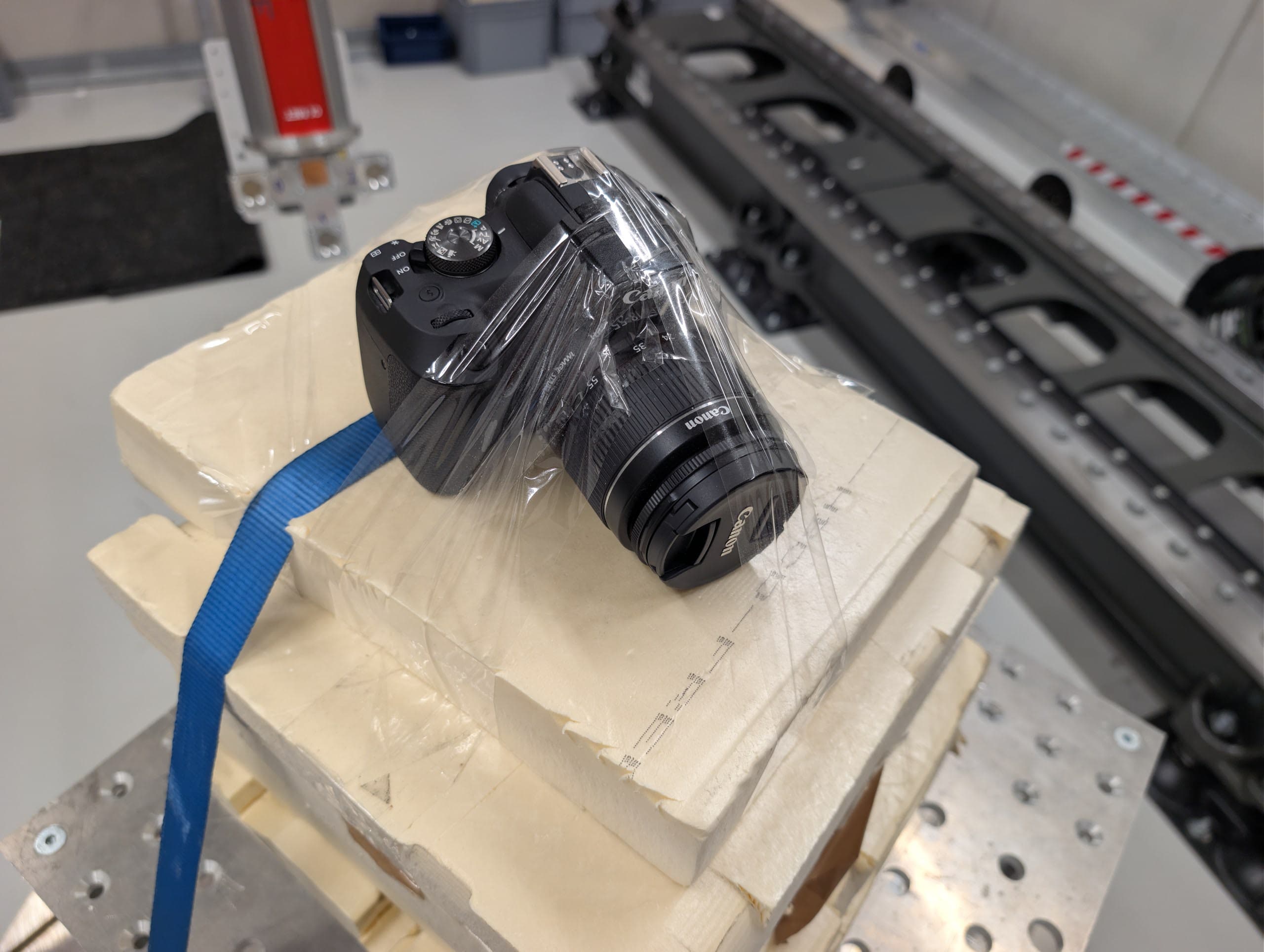}%
\label{fig:camera_sample}}
\hfill
\subfloat[]{%
\includegraphics[width=0.48\columnwidth]{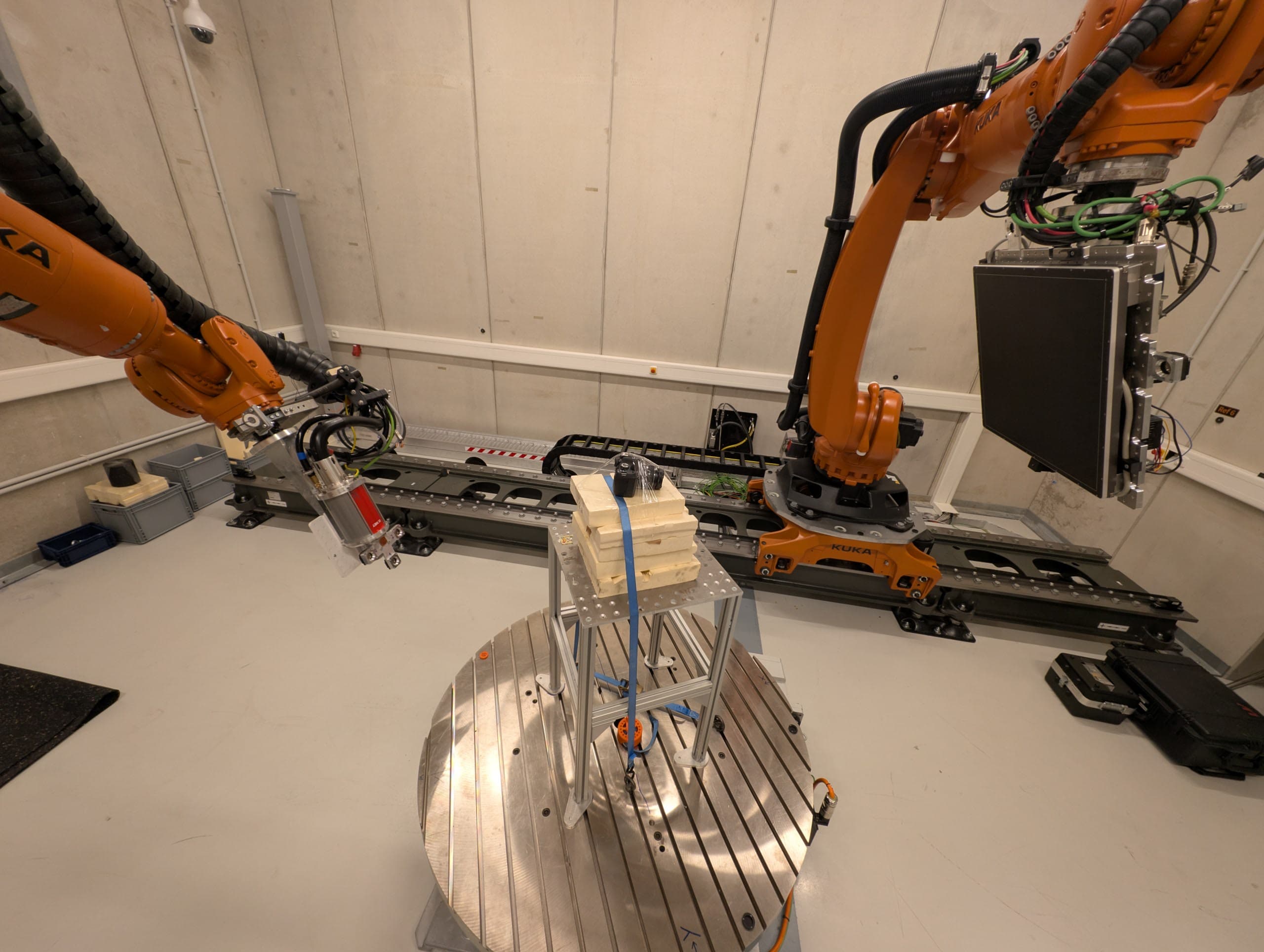}%
\label{fig:camera_bench}}
\caption{The scanned object mounted on the rotation stage (a) and the custom robot-based cone-beam CT test bench with the two-axis source and detector manipulators (b).}
\label{fig:camera_setup}
\end{figure}

The robot CT bench (\cref{fig:camera_setup}) scans a digital camera over $360^{\circ}$ azimuth and $\pm30^{\circ}$ elevation. A $120$-view circular prescan is the planning prior; a separate $1200$-view circular FDK reference is withheld until evaluation.

Coverage--bundle and full-composite pose sets are acquired at $k\in\{100,400\}$. Baselines are equidistant circular-reference subsets and uniform-on-band sets formed by matching Fibonacci targets to distinct views pooled from the four band acquisitions (mean mismatch $3$--$3.5^{\circ}$). All sparse arms use total-variation (TV) regularised adaptive steepest descent projection onto convex sets (ASD-POCS) with the TV weight chosen by data residual, without reference access. The planned arms use the historical $32$-sample bundle rule of Supplementary \cref{supp-sec:supp_camera_quadrature}. Reconstructions are intensity-matched to the reference before scoring. Supplementary \cref{supp-sec:supp_measured_reconstruction} gives acquisition and reconstruction details.

\begin{table*}[t]
\centering
\caption{Sphere-based selectors on both virtual objects, cold-started, mean $\pm$ population standard deviation. PSNR in dB and SSIM are higher-is-better, HFEN is lower-is-better. The soft-Tuy greedy row is the discrete fixed-grid selector on the same $720$-point sphere and the initialisation of every continuous row, so the three arms below it measure what leaving the candidate grid adds. Bold marks the best value per column within each object.}
\label{tab:ornl_runs}
  \scriptsize \setlength{\tabcolsep}{5pt}

\begin{tabular}{@{}l l ccc ccc@{}}
\toprule
    & & \multicolumn{3}{c}{$k = 40$} & \multicolumn{3}{c}{$k = 80$} \\
\cmidrule(lr){3-5}\cmidrule(lr){6-8}
    Object & Selector & PSNR & SSIM & HFEN & PSNR & SSIM & HFEN \\
\midrule
    \multirow{5}{*}{Defrise flange, $384^{3}$} & VCLS, discrete & $\mathbf{45.82}{\pm}0.02$ & $\mathbf{0.981}{\pm}0.000$ & $1.81{\pm}0.02$ & $49.76{\pm}0.09$ & $\mathbf{0.992}{\pm}0.000$ & $0.91{\pm}0.02$ \\
     & soft-Tuy greedy, discrete & $41.32{\pm}0.56$ & $0.951{\pm}0.003$ & $3.46{\pm}0.37$ & $45.25{\pm}0.39$ & $0.980{\pm}0.000$ & $1.87{\pm}0.15$ \\
     & cov+bundle & $44.66{\pm}0.60$ & $0.963{\pm}0.004$ & $2.00{\pm}0.15$ & $47.07{\pm}0.39$ & $0.981{\pm}0.000$ & $1.43{\pm}0.10$ \\
     & cov+VCL & $43.91{\pm}0.53$ & $0.962{\pm}0.002$ & $2.36{\pm}0.21$ & $49.36{\pm}0.15$ & $0.989{\pm}0.000$ & $0.95{\pm}0.02$ \\
     & cov+VCL+bundle & $45.72{\pm}0.25$ & $0.970{\pm}0.002$ & $\mathbf{1.78}{\pm}0.07$ & $\mathbf{50.32}{\pm}0.10$ & $0.988{\pm}0.000$ & $\mathbf{0.81}{\pm}0.00$ \\
\midrule
    \multirow{5}{*}{Fuel nozzle, $512^{3}$} & VCLS, discrete & $28.99{\pm}0.17$ & $0.770{\pm}0.004$ & $11.19{\pm}0.16$ & $34.81{\pm}0.11$ & $0.883{\pm}0.002$ & $7.94{\pm}0.11$ \\
     & soft-Tuy greedy, discrete & $29.19{\pm}0.14$ & $0.770{\pm}0.000$ & $11.33{\pm}0.19$ & $35.04{\pm}0.56$ & $0.889{\pm}0.005$ & $7.94{\pm}0.54$ \\
     & cov+bundle & $28.83{\pm}0.06$ & $0.772{\pm}0.006$ & $11.80{\pm}0.13$ & $31.44{\pm}0.06$ & $0.838{\pm}0.001$ & $9.20{\pm}0.10$ \\
     & cov+VCL & $29.94{\pm}0.49$ & $0.784{\pm}0.011$ & $10.99{\pm}0.53$ & $\mathbf{35.27}{\pm}0.24$ & $\mathbf{0.891}{\pm}0.002$ & $\mathbf{7.63}{\pm}0.17$ \\
     & cov+VCL+bundle & $\mathbf{30.08}{\pm}0.13$ & $\mathbf{0.792}{\pm}0.003$ & $\mathbf{10.55}{\pm}0.11$ & $33.19{\pm}0.34$ & $0.866{\pm}0.006$ & $7.86{\pm}0.24$ \\
\bottomrule
\end{tabular}
\end{table*}

\subsection{Compared Methods and Evaluation}
\label{sec:setup_baselines}
We compare discrete VCLS~\cite{Lin2025Tomographic}, discrete soft-Tuy greedy~\cite{Schneider2026}, and greedy-initialised continuous coverage plus VCL, bundle, or both. Discrete pools contain $720$ Fibonacci-sphere candidates, or $360$ feasible candidates in the constrained study. All virtual selectors use the same SART reconstruction; Supplementary \cref{supp-sec:supp_ablation} reports the derivative and optimiser ablation.

Metrics are peak signal-to-noise ratio (PSNR), three-dimensional structural similarity (SSIM)~\cite{Wang2004ssim}, both higher is better, and the high-frequency error norm (HFEN)~\cite{Ravishankar2011HFEN}, lower is better, with defect-resolved spherical ROIs on the flange. Uncertainties are population standard deviations over three seeds unless stated otherwise.

\subsection{Optimisation Settings and Implementation}
\label{sec:setup_hyperparams}
Virtual selection uses $z=2000$ Radon normals, $100$ Adam steps, and a cosine schedule from $\eta_0=0.05$ in the two-axis chart. Coverage has unit weight, VCL weight $0.2$ when active with $r_1=10^{-3}$, and the $5\times9$ bundle weight is $0.2/\operatorname{median}\bar\tau$. Measured planning uses $150$ steps, the constrained-elevation chart, and a $128^3$ VCL model with voxel sampling fraction $r_1=10^{-3}$ and ridge $\lambda_R=10^{-3}$ and $\varepsilon_{\mathrm{vcl}}=0.5$~mm. Implementation uses MLX~\cite{MLX2023} with Torch/CUDA projections and reconstructions on a Linux workstation with two NVIDIA RTX PRO 6000 GPUs.

\section{Results}
\label{sec:results}

\subsection{Noise-Free Benchmarks}
\label{sec:results_noisefree}
The near-axisymmetric fuel nozzle is the conservative case for a continuous selector, because a circular orbit draws an object-specific geometric advantage on this part, so we compare the sphere-based selectors among themselves. \Cref{tab:ornl_runs} reports all five families on both noise-free benchmarks.

On the nozzle the information term is what carries the continuous arms. Coverage and VCL exceed the discrete sphere baseline at both budgets, so a continuous search reaches this object without the candidate basis the swap search needs. Coverage and absorption alone fall behind that baseline, at the larger budget by more than three decibels. A near-axisymmetric part offers few geometrically distinguishable directions, so a purely geometric criterion has little to work with. Adding the information term recovers the loss at the sparse budget, where the three-term arm is the best selector on all three metrics. At the larger budget it does not, and the absorption term then costs about two decibels against coverage and VCL alone. The same arm and protocol produce the analytic column of Supplementary \cref{supp-tab:fd_vs_bundle}.

The soft-Tuy greedy row prices the off-grid step, since it is the selection every continuous arm starts from. On the flange the best continuous arm gains four decibels over it at the sparse budget and five at the larger one, cutting HFEN by about half. On the nozzle the same step buys less than a decibel, because the greedy grid solution there is already ahead of the swap search. Whether continuous refinement pays is therefore a property of the object, and it pays where the informative directions lie between the candidates.

Supplementary \cref{supp-sec:supp_ablation} compares the analytic gradient with finite differences and CMA-ES under matched objectives and evaluation budgets.

\begin{figure}[t]
\centering
\includegraphics[width=\columnwidth]{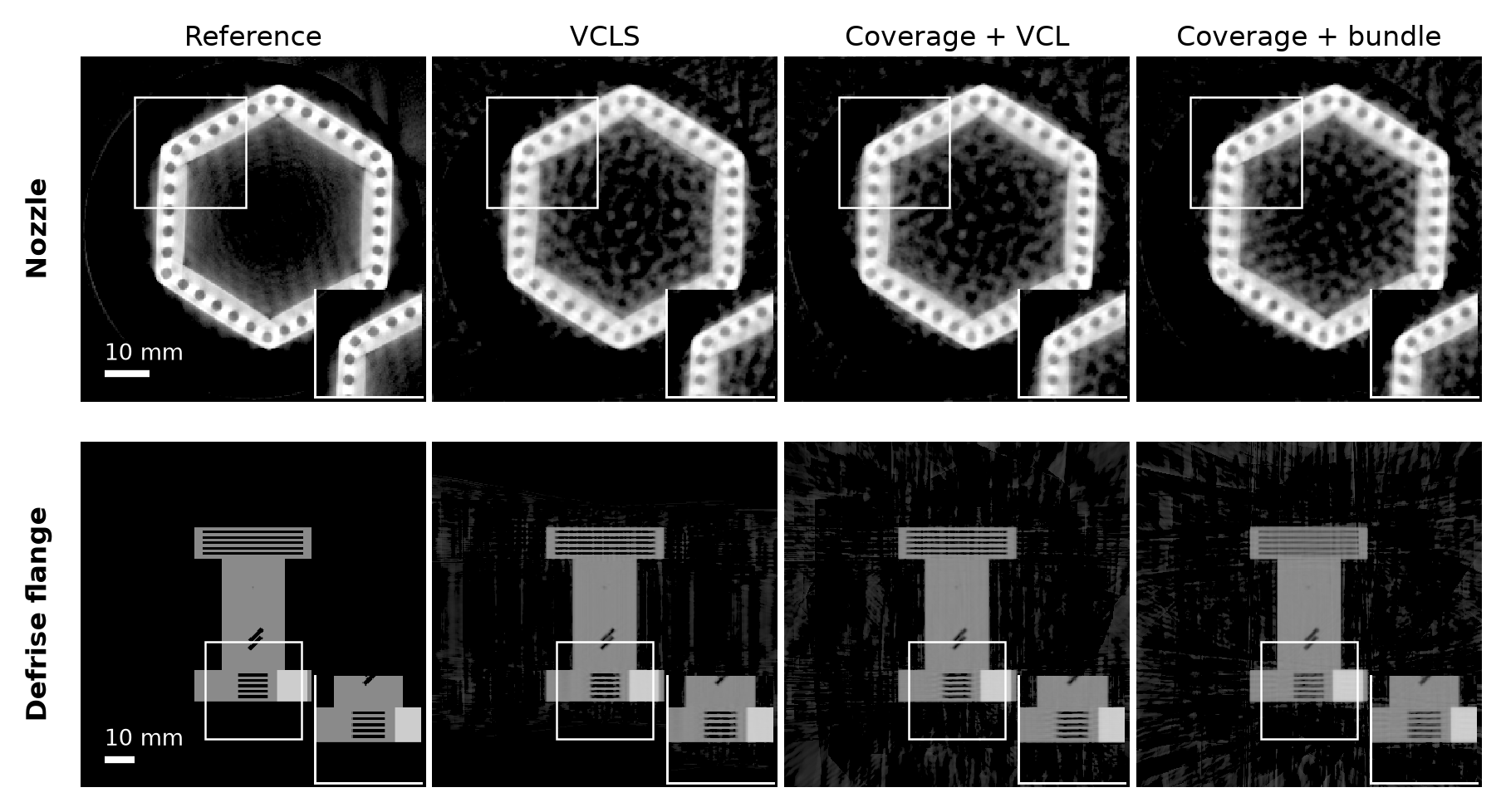}
\caption{Noise-free reconstructions at $k=40$, the nozzle rendered at $256^{3}$ for legibility. The top row is a central axial slice of the fuel nozzle with the cooling-hole ring in the inset, the bottom row a central sagittal slice of the Defrise flange with the upper lack-of-fusion stack. The nozzle selections are visually equivalent, as the sub-decibel margins of \cref{tab:ornl_runs} predict, whereas the flange separates them in the streak background.}
\label{fig:ornl_recon}
\end{figure}

Photon noise preserves the PSNR ordering of the selectors on the flange and compresses it on the dense nozzle; Supplementary \cref{supp-sec:supp_noise} reports the noise model, protocol, and full comparison.

\subsection{Constrained Acquisition}
\label{sec:results_constrained}
We evaluate the native Defrise flange at $K_{\max}=360$ and $k\in\{40,80\}$ on four reachable sets: the equatorial circle, the bench's $\pm30^{\circ}$ elevation band, a C-arm envelope with $\pm110^{\circ}$ orbital rotation and $\pm45^{\circ}$ elevation, and the full sphere. \Cref{tab:kinematics} isolates the effect of the reachable set under a fixed selector, with defect-resolved metrics.

\begin{table*}[t]
\centering
\caption{Kinematic study on the Defrise flange, noise-free, native $384^{3}$, mean $\pm$ population standard deviation. ROI-PSNR is measured at the upper lack-of-fusion stack, the designed circle-blind defect family; the lower stack behaves alike. Every row uses the same composite objective, greedy initialisation and pose parametrisation; only the reachable set changes, entering the refinement stage as bounds. Bold marks the best value per column.}
\label{tab:kinematics}
  \scriptsize \setlength{\tabcolsep}{3.2pt}
\begin{tabular}{@{}l cccc cccc@{}}
\toprule
    & \multicolumn{4}{c}{$k=40$} & \multicolumn{4}{c}{$k=80$} \\
\cmidrule(lr){2-5}\cmidrule(lr){6-9}
    Reachable set & PSNR & SSIM & HFEN & ROI-PSNR & PSNR & SSIM & HFEN & ROI-PSNR \\
\midrule
    equatorial circle & $42.56{\pm}0.00$ & $\mathbf{0.972}{\pm}0.000$ & $2.91{\pm}0.00$ & $25.36{\pm}0.00$ & $44.67{\pm}0.00$ & $0.985{\pm}0.000$ & $2.14{\pm}0.00$ & $26.00{\pm}0.00$ \\
\midrule
    band, $\pm30^{\circ}$ elevation & $43.93{\pm}0.21$ & $0.958{\pm}0.002$ & $2.26{\pm}0.09$ & $30.61{\pm}1.01$ & $48.55{\pm}0.06$ & $0.987{\pm}0.000$ & $1.11{\pm}0.02$ & $36.83{\pm}0.84$ \\
    C-arm, $\pm110^{\circ}/\pm45^{\circ}$ & $42.08{\pm}0.51$ & $0.946{\pm}0.004$ & $2.88{\pm}0.20$ & $30.45{\pm}1.34$ & $46.98{\pm}0.66$ & $0.980{\pm}0.002$ & $1.48{\pm}0.17$ & $36.91{\pm}1.70$ \\
\midrule
    full sphere & $\mathbf{45.61}{\pm}0.34$ & $0.968{\pm}0.003$ & $\mathbf{1.78}{\pm}0.08$ & $\mathbf{34.64}{\pm}0.74$ & $\mathbf{51.01}{\pm}0.28$ & $\mathbf{0.990}{\pm}0.001$ & $\mathbf{0.72}{\pm}0.04$ & $\mathbf{39.56}{\pm}0.98$ \\
\bottomrule
\end{tabular}
\end{table*}

At the LoF stacks, the equatorial circle trails the out-of-plane sets by five to nine decibels in ROI-PSNR at $k=40$ and by more than ten at $k=80$. The growing deficit indicates missing Radon data rather than undersampling; bulk-dominated global metrics understate it by at least a factor of two. Its bulk-dominated SSIM advantage persists only at the sparse budget.

\begin{figure*}[!t]
\centering
\includegraphics[width=\textwidth]{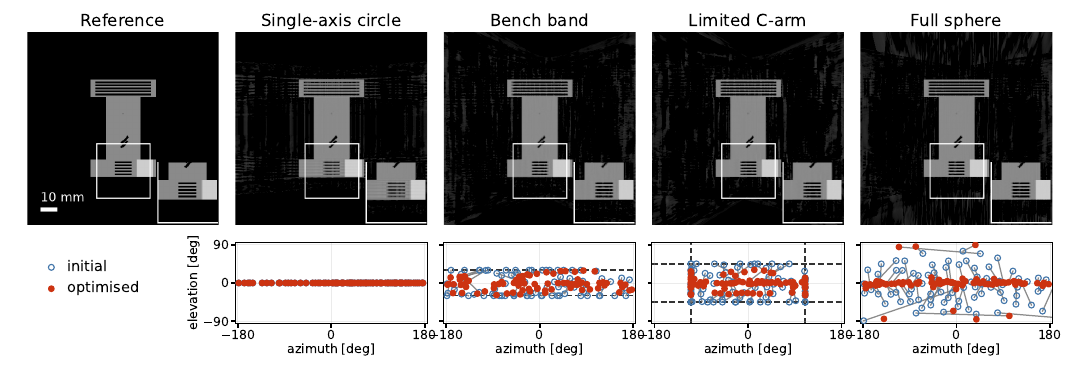}
\caption{Defrise flange at $k=80$, native $384^3$. Each geometry column pairs the sagittal reconstruction (top, with LoF inset) with its source poses (bottom). The equatorial circle loses laminar contrast; the bench band already resolves the layers. Open blue circles show Tuy-greedy poses, filled red dots the optimised poses, grey lines their displacement, and dashed lines the reachable limits. Trajectory panels share the elevation scale.}
\label{fig:kinematics_recon}
\label{fig:pose_motion}
\end{figure*}

The bench band recovers most of the full-sphere gain and matches the larger C-arm envelope at the defects. The C-arm's global PSNR nevertheless falls below the circle at $k=40$: coverage drives about a third of its poses towards the azimuth limit, concentrating viewing directions. The full sphere remains the strongest reachable set at both budgets, globally and at the defects. \Cref{fig:kinematics_recon} pairs each reconstruction with its trajectory.

Poses move off-grid by median great-circle angles of roughly one to three candidate spacings on the two-dimensional reachable sets (\cref{fig:pose_motion}); the circle remains unchanged. On the free sphere, $74\%$ of poses finish within $5^{\circ}$ of the source plane, but eleven of eighty lie at the LoF-stack heights of $20$--$27$~mm and seven remain above $60^{\circ}$ elevation. These out-of-plane poses carry the defect gain.

\subsection{Scout Robustness and Selection Cost}
\label{sec:results_planning}
We plan on $15$-iteration SART prescans of the native Defrise flange with $n_{\mathrm{s}}\in\{8,16,32,64\}$ views, then evaluate against the true volume. Circular and $\pm30^{\circ}$-band prescans use equal view budgets, isolating prescan geometry from measurement cost.

\begin{table}[t]
\centering
\caption{Prescan sensitivity on the Defrise flange at native $384^{3}$, $k=40$, $K_{\max}=360$. Each entry is the paired PSNR difference in dB between planning on a prescan reconstruction from $n_{\mathrm{s}}$ views and planning on the true volume, so zero is the oracle plan and negative is worse, mean over five seeds. Only our own selectors are listed.}
\label{tab:scout}
  \small \setlength{\tabcolsep}{6pt}
\begin{tabular}{@{}l l cccc@{}}
\toprule
& & \multicolumn{4}{c}{$n_{\mathrm{s}}$ prescan views} \\
\cmidrule(lr){3-6}
    Selector & Prescan & 8 & 16 & 32 & 64 \\
\midrule
    \multirow{2}{*}{cov+bundle} & circle & $-2.67$ & $-1.05$ & $-1.69$ & $-2.10$ \\
     & band & $\mathbf{-0.04}$ & $-0.81$ & $-1.24$ & $\mathbf{-0.02}$ \\
\midrule
    \multirow{2}{*}{cov+VCL+bundle} & circle & $-5.62$ & $-2.37$ & $-3.21$ & $-4.86$ \\
     & band & $-3.20$ & $-2.35$ & $-2.75$ & $-1.40$ \\
\bottomrule
\end{tabular}
\end{table}

\Cref{tab:scout} shows that prescan geometry matters more than budget. Circular prescans cost the three-term selector several decibels, and additional views do not reliably recover the loss. The attenuation map remains strongly correlated with the true one, but missing-cone artefacts mislead the information term and drive the planned poses to higher elevations. A band prescan at the same measurement cost removes most of the penalty at $8$ and $64$ views, less at $32$, and none at $16$. The flange is deliberately an extreme case.

\begin{table}[t]
\centering
\caption{Selection wall-clock in seconds at $K_{\max}=720$ on one Linux workstation with the Torch/CUDA backend, reconstruction excluded. VCLS is split into the object-specific candidate-basis precompute and the swap loop that uses it. Ours is the greedy-initialised coverage plus bundle arm. Best per row in bold, second best underlined.}
\label{tab:cost}
  \small \setlength{\tabcolsep}{2.6pt}
\begin{tabular}{@{}l c c rr r r r@{}}
\toprule
& & & \multicolumn{3}{c}{VCLS} & Ours & \\
\cmidrule(lr){4-6}
    Phantom & Grid & $k$ & Precomp. & Swap & Total & Total & Gain \\
\midrule
    \multirow{2}{*}{Defrise flange} & \multirow{2}{*}{$384^{3}$} & 40 & 98.1 & 0.6 & $\underline{98.7}$ & $\mathbf{5.1}$ & $19\times$ \\
     &  & 80 & 98.1 & 5.0 & $\underline{103.1}$ & $\mathbf{10.1}$ & $10\times$ \\
\midrule
    \multirow{2}{*}{Fuel nozzle} & \multirow{2}{*}{$512^{3}$} & 40 & 262.1 & 1.2 & $\underline{263.3}$ & $\mathbf{5.6}$ & $47\times$ \\
     &  & 80 & 262.1 & 3.6 & $\underline{265.8}$ & $\mathbf{13.0}$ & $20\times$ \\
\bottomrule
\end{tabular}
\end{table}

\Cref{tab:cost} prices selection itself, with reconstruction excluded and the VCLS precompute separated from the swap loop it feeds. Our objective scales mainly with the view budget, not the reconstruction grid, whereas VCLS ties its precompute to the grid and must be rebuilt for each new part. The margin therefore grows with resolution instead of shrinking.

\subsection{Measured Robot-Based CT}
\label{sec:results_real}
Planning uses only the $120$-view prescan of \cref{sec:setup_real}, and the $1200$-view reference is withheld until evaluation. The optimised poses span the available elevation band.

On the measured prescan, coverage saturates within about thirty steps. At $k=400$ the optimisation then reduces the mean bundle integral by roughly a third while coverage changes by less than one percent, so the two terms separate cleanly. This is geometric evidence only, and the acquired scans decide whether it improves image quality.

\begin{figure*}[!t]
\centering
\includegraphics[width=0.94\textwidth]{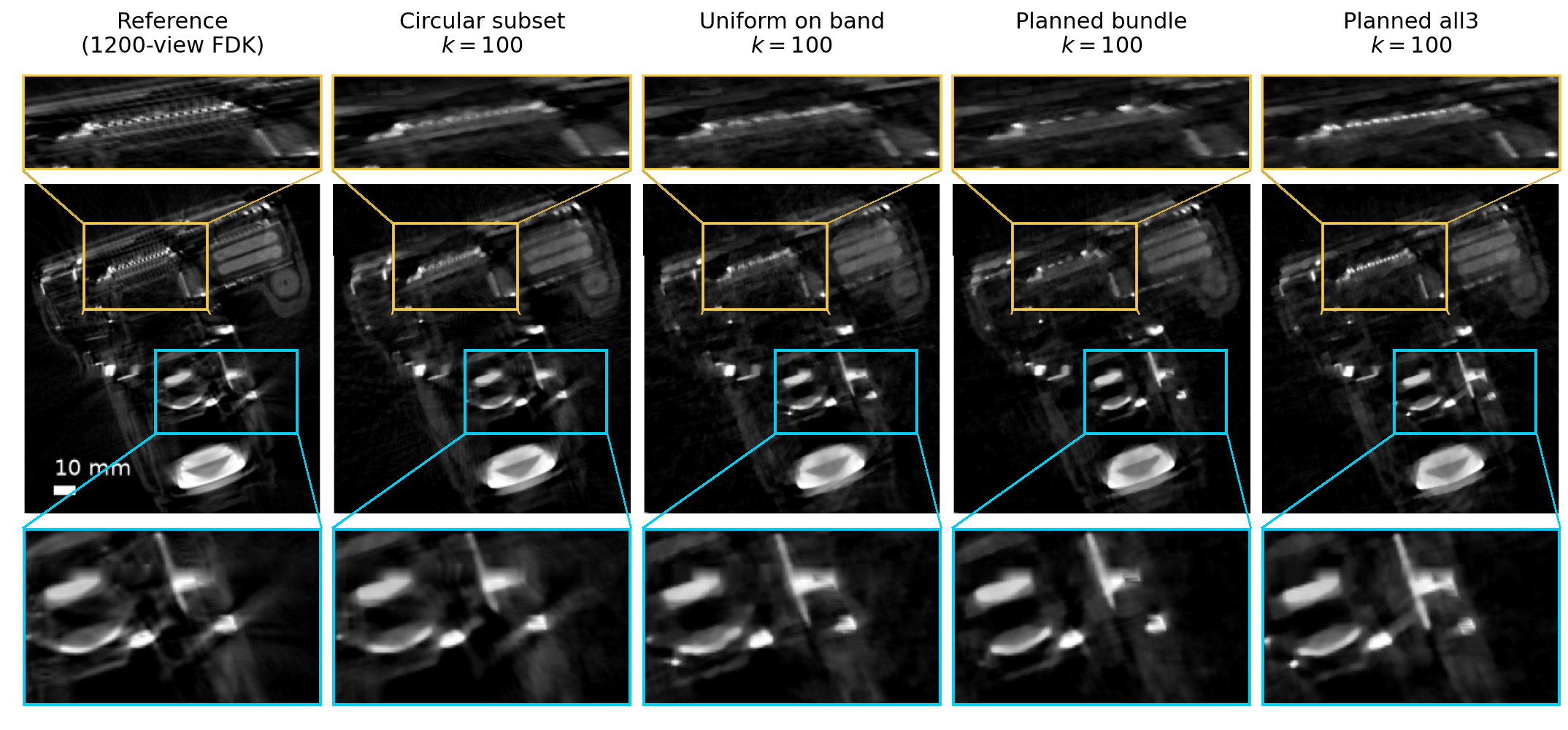}
\caption{Measured central axial slices at $k=100$, showing the quantitative $1200$-view FDK reference, the reference-coupled circular subset, same-manifold uniform sampling, and the two acquired planned arms.  Gold boxes and the strip above show the complete pin region. Cyan boxes and the strip below show a high-attenuation component above the lens together with the streaks around it. All panels use the common reference grey-value window. Band acquisitions are reconstructed natively in the reference frame, and all panels use the residual-selected TV weight.}
\label{fig:real_slices}
\end{figure*}

\begin{table}[t]
\centering
\caption{Measured camera experiment against the quantitative $1200$-view FDK reference. The circular subsets share the reference acquisition. Every metric is computed on the same reference-frame full-volume pair. Best values within each budget are bold and second-best values are underlined.}
\label{tab:real_results}
  \small \setlength{\tabcolsep}{3.5pt}
\begin{tabular}{@{}l c ccc@{}}
\toprule
    Trajectory & $k$ & PSNR & SSIM & HFEN \\
\midrule
    circular subset        & 100 & \textbf{50.02} & \textbf{0.9785} & \textbf{16.93} \\
    uniform on band        & 100 & \underline{45.53} & \underline{0.9552} & \underline{27.21} \\
    planned bundle         & 100 & 44.22 & 0.9495 & 31.31 \\
    planned full composite & 100 & 44.09 & 0.9510 & 32.95 \\
\midrule
    circular subset        & 400 & \textbf{51.49} & \textbf{0.9795} & \textbf{12.11} \\
    uniform on band        & 400 & \underline{45.09} & 0.9523 & \underline{27.58}  \\
    planned bundle         & 400 & 43.71 & 0.9485 & 32.45 \\
    planned full composite & 400 & 45.06 & \underline{0.9558} & 28.84\\
\bottomrule
\end{tabular}
\end{table}

At $k=100$, uniform sampling leads both planned arms on every metric; at $k=400$ it leads the bundle-only arm throughout and the three-term arm on two of three metrics. The reference-based metrics therefore show no consistent gain over uniform sampling on the same band, though the imperfect reference discussed below warrants caution in reading them. The visual comparison in~\cref{fig:real_slices} adds a distinction the aggregate numbers miss; the three-term arm reconstructs the pins in the upper enlargement most completely, at the shown budget and at $k=400$ alike, whereas the bundle-only arm yields a clean overall slice whose fine detail remains imperfect in places.

The circular subset is best on every metric, but is not a clean selector comparison because it is drawn directly from the reference acquisition, avoids cross-session alignment, and does not use the $\pm30^{\circ}$ elevation range. The reference also contains radial streaks that are strongest for the circular subset and progressively weaker for uniform band sampling and the planned arms, consistent with suppression by out-of-plane views. Since the metrics reward agreement with this artefact-bearing reference, they favour the circular subset and do not reflect absolute fidelity. We therefore rely primarily on fixed-budget comparisons among the separately acquired band pose sets.

\section{Discussion}
\label{sec:discussion}
The matched benchmarks show that continuous selection reaches the quality of discrete swap search. VCL approximates normalised mean-squared error~\cite{Lin2025Tomographic}, so a strong VCL solution leaves little PSNR headroom. Off-grid refinement adds four to five decibels on the asymmetric flange but less than one on the near-axisymmetric nozzle. The estimator ablation attributes final quality to the objective and selection cost to the gradient; photon noise compresses the margins.

The reachable set outweighs the selector. Leaving the equatorial plane recovers the designed laminae, and the moderate bench band captures most of the free-sphere gain. On bounded envelopes, however, coverage drives poses towards the azimuth limit and lowers global performance relative to the full sphere. Planning-prior geometry matters for the same reason: circular-prescan artefacts mislead the continuous information term, while distributing the same prescan views over the reachable band removes most of the penalty at most budgets. Selection cost follows the view budget rather than the reconstruction grid, so the advantage over the discrete precompute grows with resolution.

The measured study demonstrates physical execution and reconstruction in a common frame, but no consistent metric gain over uniform band sampling. The circular reference contains streaks suppressed by the planned arms, so reference-based metrics partly measure artefact agreement. The three-term arm resolves the fine pins most completely in \cref{fig:real_slices}.

Geometry-only coverage is the default without a prior or under tight selection cost; the bundle term is preferable when a prior exists and attenuation or occlusion matters. Attenuation can also pull poses into unfavourable geometry, as on the Defrise phantom, where adding the information term keeps poses out of plane.

Five limitations bound these claims. The optimiser is local and the objective is degenerate at these budgets. Photon noise is modelled on the virtual benchmarks only, as monochromatic Poisson transmission noise at one idealised incident fluence, without scatter, detector effects, beam hardening or a calibrated polychromatic spectrum. The measured comparison confounds several effects, since the circular subsets share the reference acquisition and the band budgets come from independently acquired sessions. The measured study plans with the historical bundle estimator. Finally, every study plans for the object that is subsequently scanned, so what the prescan study bounds is the sparsity of the prior and not a mismatch between prior and part.

Two directions follow. Ordering the poses and interpolating them into a mechanically smooth path remains a subsequent step; optimising the coefficients of a trajectory parametrisation instead, for instance within the space-filling source paths of Kingston et al.~\cite{Kingston2018}, would impose continuity and velocity constraints directly and is also the most promising route to reducing the initialisation dependence. Differentiating through a limited number of reconstruction iterations would admit task-specific image losses and opens the way to replanning the remaining poses from the projections already measured, so that the planning prior is progressively replaced by data from the object itself.

\section{Conclusion}
\label{sec:conclusion}
Optimising source poses on the scanner's kinematic manifold provides a portable alternative to fixed candidate pools and their object-specific precomputation. The continuous selector reaches discrete-search quality on the free sphere while incorporating attenuation and scanner limits directly into selection. Reachable geometry and the planning prior govern the remaining differences. Continuous planning is therefore useful when per-part precomputation cannot be amortised, informative directions lie between candidates, or scanner kinematics must enter the objective. The robot study establishes acquisition feasibility, while consistent measured image-quality gains remain to be demonstrated.

\section*{Code and Data Availability}
\label{sec:availability}
The implementation, tests, paper configurations, and rerun manifest are available in
\href{https://github.com/Linda-SophieSchneider/Gradient-Based-Trajectory-Optimisation-over-Continuous-Poses-for-Sparse-View-Cone-Beam-CT}{the authors' GitHub repository}.
The ORNL fuel-nozzle volume is available through the release cited
in~\cite{Ziabari2025ORNLData}. Access to the phantom data can be requested from the
corresponding author, subject to the source data and institutional
redistribution conditions.

\section*{Acknowledgment}
\label{sec:ack}
The authors used Claude, a large language model by Anthropic~\cite{Anthropic2025Claude}, for textual refinement and shortening of the manuscript and for correction and structuring of the research code. The model produced no experimental results, no measured numbers, and no scientific claims. The authors reviewed and verified all content and take full responsibility for it.

\bibliographystyle{IEEEtran}
\bibliography{content/ppg_ref}
\end{document}